\documentclass{ieeeaccess}

\usepackage{verbatim}

\usepackage[caption=false,font=normalsize,labelfont=sf,textfont=sf]{subfig}

\usepackage{cite}
\usepackage{amsmath,amssymb,amsfonts}
\usepackage{algorithmic}
\usepackage{graphicx}
\usepackage{caption}
\usepackage{adjustbox}
\usepackage{tabularx}
\usepackage{breakurl}
\usepackage{longtable}
\usepackage{array}

\usepackage[breaklinks=true]{hyperref}
\usepackage{cite}
\usepackage{textcomp}
\usepackage[table]{xcolor}
\usepackage{enumitem}
\usepackage{multicol}
\usepackage{multirow}
\usepackage{xcolor}

\newcommand\subash[1]{\textcolor{red}{Subash: #1}}
\newcommand\ivan[1]{\textcolor{orange}{Ivan: #1}}

\usepackage{booktabs,tabularx,enumitem,ragged2e}

\usepackage{colortbl}

\def\BibTeX{{\rm B\kern-.05em{\sc i\kern-.025em b}\kern-.08em
    T\kern-.1667em\lower.7ex\hbox{E}\kern-.125emX}}
\begin{document}
\doi{}

\title{Security Considerations in AI-Robotics: A Survey of Current Methods, Challenges, and Opportunities}
%

\author{\uppercase{Subash Neupane\authorrefmark{1}}, 
\uppercase{Shaswata Mitra\authorrefmark{2}}, 
\uppercase{Ivan A. Fernandez\authorrefmark{3}}, 
\uppercase{Swayamjit Saha\authorrefmark{4}}, 
\uppercase{Sudip Mittal\authorrefmark{5}} , 
\uppercase{Jingdao Chen\authorrefmark{6}} , 
\uppercase{Nisha Pillai\authorrefmark{7}} ,
\uppercase{Shahram Rahimi\authorrefmark{8}} }

\address[1]{Department of Computer Science \& Engineering, Mississippi State University, Mississippi, (email: sn922@msstate.edu)}
\address[2]{Department of Computer Science \& Engineering, Mississippi State University, Mississippi, (email: sm3843@msstate.edu)}
\address[3]{Department of Computer Science \& Engineering, Mississippi State University, Mississippi, (email: iaf28@msstate.edu)}
\address[4]{Department of Computer Science \& Engineering, Mississippi State University, Mississippi, (email: ss4706@msstate.edu)}
\address[5]{Department of Computer Science \& Engineering, Mississippi State University, Mississippi, (email: mittal@cse.msstate.edu)}
\address[6]{Department of Computer Science \& Engineering, Mississippi State University, Mississippi, (email: chenjingdao@cse.msstate.edu)}
\address[7]{Department of Computer Science \& Engineering, Mississippi State University, Mississippi, (email: pillai@cse.msstate.edu)}
\address[8]{Department of Computer Science \& Engineering, Mississippi State University, Mississippi, (email: rahimi@cse.msstate.edu)}

\tfootnote{}

\markboth
{Security Considerations in AI-Robotics}
{Neupane \headeretal}

\corresp{Corresponding author: Subash Neupane (e-mail: sn922@msstate.edu)}


\begin{abstract}

Robotics and Artificial Intelligence (AI) have been inextricably intertwined since their inception. Today, AI-Robotics systems have become an integral part of our daily lives, from robotic vacuum cleaners to semi-autonomous cars. These systems are built upon three fundamental architectural elements: \textit{perception, navigation and planning}, and \textit{control}. However, while the integration of AI in Robotics systems has enhanced the quality of our lives, it has also presented a serious problem - these systems are vulnerable to security attacks. The physical components, algorithms, and data that makeup AI-Robotics systems can be exploited by malicious actors, potentially leading to dire consequences. Motivated by the need to address the security concerns in AI-Robotics systems, this paper presents a comprehensive survey and taxonomy across three dimensions: \textit{attack surfaces, ethical and legal concerns}, and \textit{Human-Robot Interaction (HRI) security}. Our goal is to provide readers, developers and other stakeholders with a holistic understanding of these areas to enhance the overall AI-Robotics system security. We begin by identifying potential attack surfaces and provide mitigating defensive strategies. We then delve into ethical issues, such as dependency and psychological impact, as well as the legal concerns regarding accountability for these systems. Besides, emerging trends such as HRI are discussed, considering privacy, integrity, safety, trustworthiness, and explainability concerns. Finally, we present our vision for future research directions in this dynamic and promising field.

\end{abstract}



\maketitle

\IEEEdisplaynontitleabstractindextext

%
\IEEEpeerreviewmaketitle

\section{Introduction and Motivation}



Robotics is a multidisciplinary field of engineering that deals with the design, construction, and operation of intelligent machines called robots, capable of sensing the environment and executing tasks autonomously or through human guidance. Robotic systems are comprised of sensors and actuators that can interact with the environment to accomplish a set of tasks \cite{robotarchi}. The field has made remarkable strides in reshaping industries such as transportation, manufacturing, healthcare, agriculture, entertainment, space exploration, military, and beyond. On the other hand, Artificial Intelligence (AI) is the simulation of human intelligence processes by machines, especially computer systems. In this context, AI-Robotics systems may be referred to as {integration and collaboration between AI and robotic technologies}. These systems combine AI algorithms, which enable machines to perform tasks that typically require human intelligence, with robotic platforms designed to interact with and manipulate the physical world. The recent advances in Machine Learning (ML) and Deep Learning (DL) techniques have paved the way for the development of robots with cognitive abilities to execute functions such as \textit{perception, navigation and planning}, and \textit{control}. This empowerment enables robots to perceive, learn, deduce, and make real-time decisions, closely mimicking human behaviors. With ML/DL algorithms continuing to advance, the synergy with robotics propels us toward a future where robots play pivotal roles in {societies}, offering unprecedented levels of automation, efficiency, and adaptability.

As AI-Robotics systems become increasingly sophisticated, versatile, and integrated into our daily lives, it also revolutionize how we work, live, and explore the world. For example, in the domain of \textit{autonomous vehicles and transportation},  self-driving cars \cite{waymo}, drones, and Unmanned Aerial Vehicles (UAVs) are revolutionizing transportation systems. Similarly, in \textit{industrial automation and 
manufacturing processes}, to improve efficiency, productivity, and safety— collaborative robots \cite{sherwani2020collaborative} perform tasks such as assembly, welding, material handling, and quality inspection. In \textit{healthcare}, AI-Robotics systems have also demonstrated their potential in revolutionizing surgery as evidenced by Smart Tissue Autonomous Robot (STAR) \cite{leonard2014smart}, and rehabilitation processes as discussed in \cite{ai2021machine}. In \textit{agriculture}, where they assist in diverse tasks like planting, harvesting, and crop health monitoring, enabling the realization of precision agriculture \cite{r2018research, gupta2020security,sontowski2020cyber}. Alongside day-to-day civilian activities, defense and military operations are equally utilizing AI-Robotics systems to improve strategic capabilities while minimizing human exposure to danger \cite{lin2008autonomous}. From autonomous surveillance drones that enhance situational awareness and warfare missions to robotic vehicles for logistics transportation and reconnaissance in hazardous environments, AI-Robotics is revolutionizing how military forces plan, execute, and adapt.

The security of robotic systems has been an active research paradigm for a long time \cite{yaacoub2022robotics, jaitly2017security, wu2007survey, mekdad2023survey, placed2023survey, li2020survey, Lamssaggad2021ASO, narayanan2016obd_securealert, chukkapalli2020ontologies}. {Due to their pervasive nature, AI-Robotics systems are susceptible to diverse security attacks by threat actors aiming to exploit vulnerabilities in the underlying AI technologies. These attacks are capable of compromising the functionality, safety, and reliability of a system.} Adversarial attacks, data breaches, and system manipulations can lead AI-Robotics systems to cause accidents, property damage, and even endanger human lives. 

{Recently, there have been several examples of adversarial attacks on semi-autonomous vehicles that demonstrate the security risk in AI-Robotics.} 
{In 2020, researchers} manipulated Tesla cars into speeding up by 50 miles per hour by subtly altering a speed limit sign on the side of a road, fooling the car’s Mobileye EyeQ3 camera system \cite{tesla_crash}. While, in 2015, a Jeep Cherokee got remotely hacked while being driven, causing Fiat Chrysler to recall 1.4 million vehicles and pay \$105 million in fines to the National Highway Traffic and Safety Administration \cite{tcsession_mobility2022}. The vulnerability occurred due to a lack of code signing in the head unit of the car or Electronic Control Unit (ECU), allowing the gateway chip to get reprogrammed and send instructions to vehicle components, such as the steering wheel and brakes \cite{jeep_cherokee}. Furthermore, attacks on \textit{confidentiality, privacy, integrity}, and \textit{availability} can affect trust and acceptance among users \cite{yaacoub2022robotics, us_ai_report, mittal2023ai,bertino2022security}. Therefore, understanding the security dimension of AI-Robotics systems and familiarizing oneself with the existing methods to defend against in the event of an attack holds the utmost importance. 

Furthermore, the integration of robots into various aspects of our lives has given rise to pressing \textit{ethical and legal} concerns. These concerns extend beyond security attacks and are intertwined with the growing cognitive capabilities of AI-Robotics systems. Ethically, the influence of robots on human behavior and emotions is a significant issue. In scenarios involving medical, social, and assistive robots, the potential for robots to form emotional bonds with users raises questions about the risk of dependence on machines and their psychological impact on individuals \cite{riek2014code}. Also, the overt reliance on autonomous robotic agents for decision-making processes can have negative impacts on human cognitive thinking and reasoning capacities, ultimately leading to a loss of human autonomy \cite{datteri2009ethical}. On the legal front, navigating liability and responsibility within robotic systems presents a complex challenge. For instance, determining accountability when a ``remote-assisted telesurgery’’ procedure goes awry requires careful consideration. Moreover, the exchange of personal data and information between robots and humans raises data privacy concerns, necessitating adherence to legal frameworks like the 2020 California Consumer Privacy Act (CCPA) \cite{CCPA} and the 2016 General Data Protection Regulation (GDPR)\cite{gdpr}.

Besides, growing cognitive capabilities of AI-Robotics systems have allowed robots to have direct cognitive interactions with humans— raising security concerns over Human-Robot Interaction (HRI). The ethical and legal dimensions of these interactions revolve around integrity, privacy, and safety. Addressing these ethical and legal intricacies is paramount to ensuring responsible, respectful, and secure interactions between robots and humans.

In contrast to prior surveys that focus on specific security aspects in AI-Robotic systems, ranging from attacks on operating systems, networks, and the physical layer \cite{botta2023cyber}, information faults \cite{dutta2021cybersecurity}, to ensuring secure communication in IoTs \cite{ dutta2021cybersecurity, alsamhi2019survey}, and securing trust between operators and cobots in automation and construction domains \cite{emaminejad2022trustworthy}, our paper sets itself apart by seamlessly integrating \textit{attack surfaces and mitigation strategies, ethical and legal concerns,} and \textit{HRI security} dimensions, providing a comprehensive review across these critical areas. {On a taxonomic front, Jahan et al.,} \cite{jahan2019security} {employs a classification approach akin to our paper, albeit with a focus solely on autonomous systems.}   

To the best of our knowledge, this paper is the first to dissect the AI-enabled attack surfaces in 
{a generic hybrid AI-Robotics architecture \ref{fig:generic}.}
{ This unique perspective enhances the overall security understanding of AI-robotics systems.}
Furthermore, we also investigate the ethical and legal concerns surrounding the accessibility, fairness, trustworthiness, explainability, and accountability of robotic applications, along with the security studies of HRI. 

The major contributions of this paper are as follows:
\begin{itemize}

    \item 
    {We present a taxonomy structured around three dimensions of AI-Robotics systems security: \textit{attack surfaces and mitigation strategies, ethical and legal concerns}, and \textit{security of HRI}. This taxonomy is designed to augment the overall security of these systems by offering stakeholders comprehensive insights across these dimensions.
}
    \item 
    {We provide a comprehensive survey of the attack surfaces within hybrid AI-robotics systems, encompassing architectural layers such as \textit{perception, navigation and planning}, and \textit{control}. In addition, we present state-of-the-art defenses aimed at significantly reducing the risk of attacks across these layers.}
    
    \item We discuss the challenges inherent within AI models, frameworks, software and explain how the existing vulnerabilities within these tools can inadvertently contribute to undermining the integrity of the final robotic systems.
    \item We present the multifaceted exploration of the ethical and legal dimensions associated with AI-Robotics systems.   
    \item We address the intricate security issues pertaining to HRI, with the focus on privacy concerns. Within this context, we recommend strategic defensive solutions.
    \item 
    {We provide research recommendations organized according to the three dimensions mentioned earlier, with the goal of enhancing the robustness and resilience of future AI-Robotic systems.} 
\end{itemize}

\noindent The remainder of the sections are as follows: Section \ref{section_architecture} provides background on robotics architecture. Section \ref{section_taxonomy} summarizes our survey and taxonomy.  
Following that, we investigate on attack surfaces on various layer of robotic architecture in Section \ref{section_attack_surfaces}. Then, we discuss ethical and legal concerns surrounding the use of AI-Robotic systems in Section \ref{section_ethical_and_legal}. Section \ref{section_HRI} discusses security studies pertaining human robot interaction. Section \ref{section_research_direction} identifies research challenges and provides future direction in developing secured robotic systems. Finally, Section \ref{section_conclusion} concludes our survey.

\section{AI-Robotics Systems Architecture}
\label{section_architecture}


\begin{figure*}[ht]
  \centering
  \includegraphics[width=\linewidth]{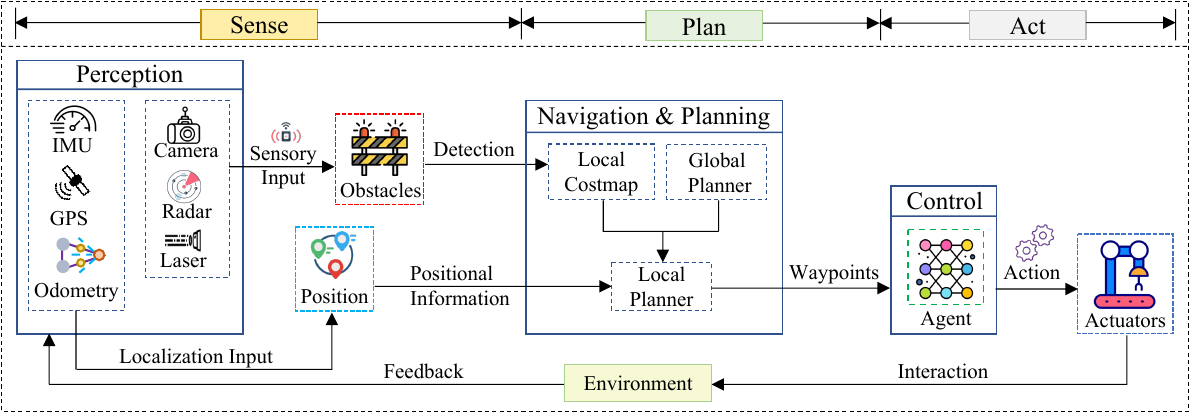}
  \caption{A generic robot architecture consisting of perception, navigation, and control stacks. Icons from \cite{flaticon}}
  \label{fig:generic}
\end{figure*}

{Robotic systems can be designed using a variety of architectural paradigms. In this section, we will explore these paradigms further.} Murphy et al. \cite{murphy2019introduction} describe a range of current architectures such as \textit{hierarchical, reactive, and hybrid deliberative-reactive} in terms of the relationships between three primitives (i.e., sense, plan, and act) for organizing intelligence in robots. Table \ref{table:robot_primitives} attempts to characterize these three fundamental building blocks based on their inputs and outputs. Hierarchical paradigms in robotics follow a top-down approach, where the robot first senses the world, then plans its next action and finally executes it. A hierarchical paradigm is also known as a deliberative architecture. Each layer in the hierarchy depends on the layer below it for information and guidance. Reactive paradigms prioritize immediate responses directly connected to sensor input, without extensive planning. Hybrid deliberative-reactive paradigms strike a balance between reactive responses and higher-level decision-making, utilizing both immediate reactions and planned actions based on sensory information.

\begin{table}[ht]
\caption{{Robot primitives defined in terms of input and output. Adapted from} \cite{murphy2019introduction}.}
{\renewcommand{\arraystretch}{1.20}%
\begin{tabular}{p{1.2cm}|p{2.3cm}|p{4.0cm}}
\hline
\rowcolor{cyan!20!}
Primitives         & Input                  & Output \\                                                                                          \hline

\multirow{3}{*}{Sense} & Sensor data from LiDAR, Cameras, GPS etc. &Sensed information such as Obstacle location, Occupancy grids etc.\\ \hline


Plan & Map information & Waypoints, Path, Trajectory  \\ \hline


Act & Waypoints & Actuator commands\\ \hline

                                    
\end{tabular}}
\\
\\

\label{table:robot_primitives}
\vspace{-2mm}

\end{table}
Modern AI-Robotics systems commonly adopt a hybrid architecture that combines both deliberative and reactive modules. Figure \ref{fig:generic} illustrates a generic three-layered hybrid architecture that corresponds to the sense-plan-act paradigm, with \textit{Perception}, \textit{Navigation \& Planning}, and \textit{Control} as its constituent layers. In the subsequent subsection, we will provide a detailed explanation of each of these layers. 

\subsection{Robotic Perception}\label{section_perception}

The perception stack of the robot architecture involves sensing and understanding relevant information about the state of the operating environment by leveraging several sensors, for example, GPS, accelerometer, inertial measurement unit (IMU), gyroscope, magnetometer, camera, and tactile sensors among others \cite{kappassov2015tactile}. The combination of multiple sensors as depicted in Fig \ref{fig:generic} make up perception stack. These sensors play a pivotal role in actively sensing the raw data that represent the physical state of 
{robotics} systems in the environment. Furthermore, this sensory data 
aids in planning and calculation of actuator signals in the 
{subsequent stages.} \cite{dash2021stealthy}.

Modern 
{AI-Robotics systems} require high-level 
{sensing} abilities, such as object detection, localization, positional information, and interaction with humans. An important aspect of achieving successful perception functions in robotics is gaining a deep understanding of the operating environment. One highly influential approach for environment understanding, especially in the context of mobile robots and autonomous vehicles, is the representation of the environment through 2D mappings known as occupancy grid mapping \cite{moravec1985high}. However, attention has now shifted towards 3D representations. Recent technological advancements, including Octomaps \cite{hornung2013efficient}, and the widespread use of RGBD sensors with depth capabilities have significantly enhanced the construction of larger and more detailed 3D maps for detailed 3D-like environment representations. Efforts have been focused on semantically labeling these maps by integrating them with Simultaneous Localization and Mapping (SLAM) frameworks. This integration enables autonomous navigation and precise self-localization based on the generated map.


Furthermore, the advances in AI have enabled the gathering, analysis, and prediction of visual, audio, depth, and tactile information that is required for perception functions \cite{karoly2020deep}. Object detection and localization are mostly performed through visual information (image) and depth information using Convolution Neural Network (CNN) architecture \cite{schwarz2015rgb, girshick2014rich, 6338939}, and its variants Recurrent CNN \cite{pinheiro2014recurrent, moore2023ura},  and convolutional auto-encoder \cite{kehl2016deep}. More recently, the state-of-the-art Visual Transformer \cite{dosovitskiy2020image} and its variants have also been used to improve perception performance \cite{xu2022v2x}, object detection, and tracking \cite{fu2022local} among others. 




\subsection{Robotic Planning and Navigation}
\label{archi-section_plan_and_nav}

The input to the navigation stack is the sensory information and map obtained by the various sensors from the perception stack [\ref{section_perception}] 
{as depicted in Fig} \ref{fig:generic}. Navigation algorithms are responsible for the generation of a series of waypoints, which are then passed on to the control stack [\ref{archi-control}]. Classic robotic navigation and planning algorithms typically involve explicit rule-based methods and predefined algorithms for path planning and obstacle avoidance. These algorithms often rely on geometric models and predefined maps of the environment. They can be categorized into four groups, including \emph{graph search algorithms} (for example, A* \cite{hart1968formal}), \emph{sampling-based algorithms} such as Rapidly-exploring Random Tree (RRT \cite{lavalle1998rapidly}), \emph{interpolating curve algorithms} \cite{reeds1990optimal}, and \emph{reaction-based algorithms} such as Dynamic Window Approach (DWA \cite{fox1997dynamic}).

AI/ML algorithms, on the other hand, offer a distinct advantage in object and environment identification as they can autonomously learn patterns and features directly from sensor data, eliminating the need for manual feature engineering or predefined models. This capability has resulted in remarkable enhancements in the performance of robotic navigation and planning, leading to its adoption for generating local costmaps as well as for both global and local planning tasks, thereby paving the way for more adaptive and efficient robotic systems.

The process of navigating towards a long-term goal is referred to as global planning. Local and global planners work together to achieve navigation goals. Using Deep Reinforcement Learning (DRL) technique, Kastner et al. \cite{kastner2021connecting}, for example, combine two planners that operate at different scales: a global planner that optimizes long-term navigation by producing high-level waypoints and a local planner that navigates from one-waypoint to the next while avoiding obstacles. 
{On the other hand}, local path planning and costmap generation refers to the process of short-term path planning to avoid obstacles and is used in mobile robots, where the navigation is solely based on local landmarks and environmental factors. Costmaps are utilized in objection tasks in small-scale robotic applications. Work that leverages CNN and costmaps can be found in \cite{perez2018learning}, while map-based Deep Reinforcement Learning (DRL) for mobile robot navigation for the obstacle detection task based on a generated local probabilistic costmap is presented in \cite{chen2020robot}.
Deep learning models that employ CNNs for local path planning by circumventing objects for autonomous navigation can be found in the works of \cite{puthussery2017deep}. Similarly, Ritcher et al. \cite{richter2017safe}, proposed a combination of a fully connected feed-forward network and an autoencoder to model collision probability, and more recently, Regler et al. \cite{regier2020deep} presented a DRL-based local planner with a conventional navigation stack, and Chen et al. \cite{chen2022deep} presented DRL for path planning, dynamic obstacle avoidance, and goal approaching. 
However, an 
{AI-Robotic} systems should also employ navigation algorithms and planning that are responsive to unexpected changes such as grasping novel objects \cite{morrison2018closing} and moving objects \cite{chen2022deep} in the dynamic environment.



\subsection{Robotic Control}
\label{archi-control}



{The input to the control stack is the waypoint information from the navigation and planning stack \ref{archi-section_plan_and_nav}, as illustrated in Fig. \ref{fig:generic}. The control stack is responsible for generating control signals by leveraging various algorithms to manipulate the behavior of robots. }It involves designing control systems that enable robots to perform specific tasks such as moving actuators based on the planned path given by the planning and navigation stack. The goal of this is to develop algorithms that can effectively and efficiently command the robot's actuators based on sensor feedback.

Some of the classic control algorithms include Proportional-Integral-Derivative (PID) control \cite{shaw2003pid} and Linear-Quadratic-Regulator (LQR) \cite{prasad2014optimal}. PID adjusts control signals based on proportional, integral, and derivative terms. It provides stable and robust control by continuously comparing desired and actual states and minimizing the error. LQR on the other hand, is used to control non-linear dynamical systems by computing the optimal control signal needed to move the system to a desired state. These algorithms, while effective for many applications, often rely on explicit mathematical models and predefined control strategies.

In recent years, there has been a significant shift towards leveraging AI to enhance robotic control capabilities. In particular, 
{Deep Reinforcement Learning} (DRL), has gained considerable attention in robotic control, including locomotion, manipulation, and autonomous vehicle control \cite{kober2013reinforcement}. DRL algorithms allow robots to learn control policies through interactions with the environment, receiving rewards or penalties based on their actions. Kalashnikov et al. \cite{kalashnikov2018scalable} demonstrated the use of DRL for controlling a robotic arm to grasp and manipulate objects with a high success rates in their study. In another line of work, the authors in \cite{sergey} leveraged DRL techniques for robot controllers to learn a range of dynamic manipulation behaviors, such as stacking large Lego blocks at a range of locations, threading wooden rings onto a tight-fitting peg, inserting a shoe tree into a shoe, screwing bottle caps onto bottles, and assembling wheels in a toy airplane. 

{In the next section, we discuss our survey approach and taxonomy.}

\section{Survey Approach and Taxonomy} \label{section_taxonomy}
\label{survey_approach}


This survey builds upon the intricate relationship between physical and digital vulnerabilities 
{within} AI-Robotic systems. 
{It} adopts a comprehensive approach by providing a taxonomy that explores the existing attack surfaces along with 
defensive strategies that are relevant to 
{AI-Robotics} systems. The primary objective is to understand how 
{various} attack vectors impact different components of robotic architecture. To achieve this, we first introduce the hybrid AI-Robotics architecture, comprising three layers: \textit{perception, navigation and planning}, and \textit{control} (see Section \ref{section_architecture}). The foundational understanding of different architectural layers of robotic systems allows us to identify the attack surfaces and their 
{impact} on these components. 

\begin{figure*}[t]
  \centering
  \includegraphics[width=\linewidth]{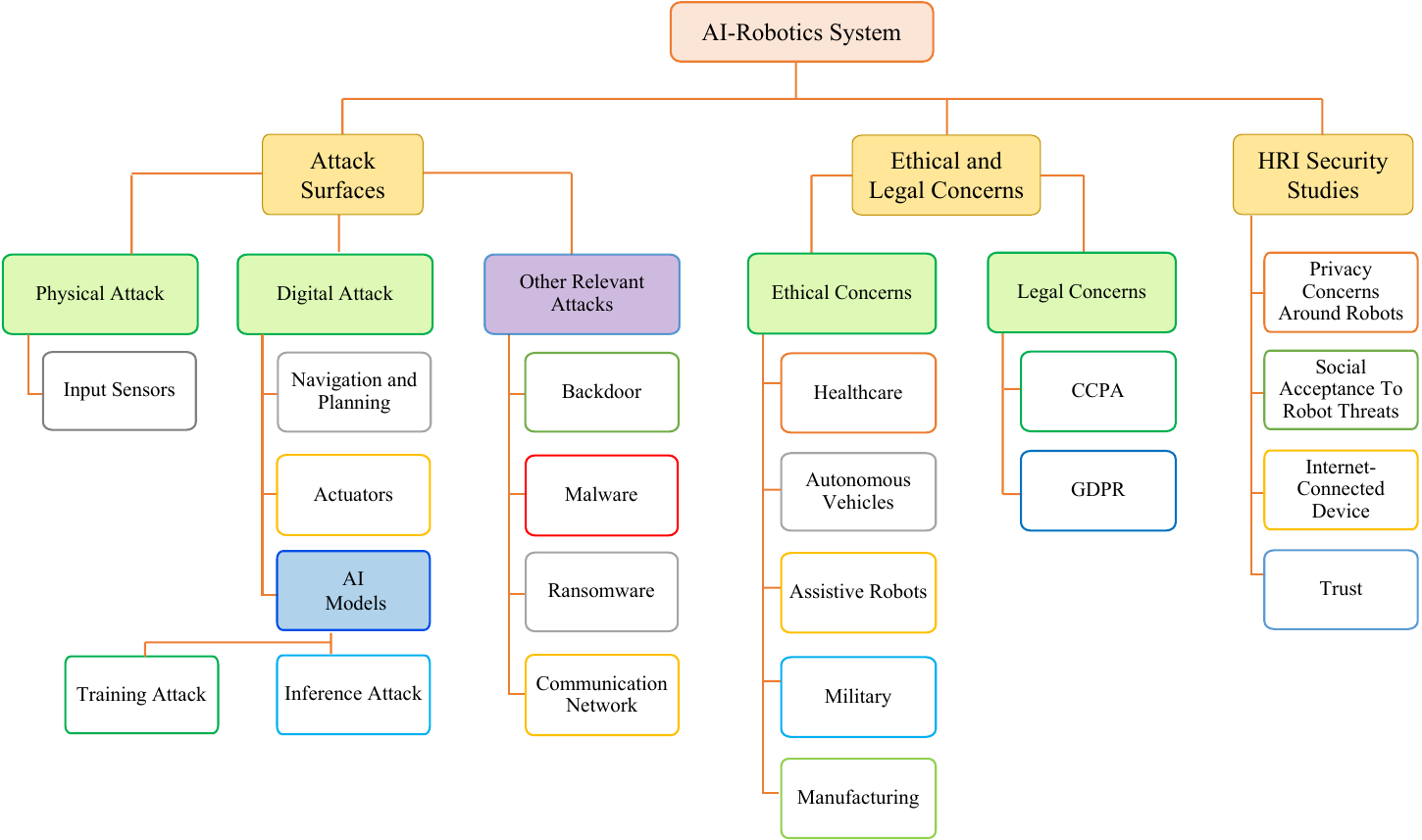}
  \caption{An overview of our proposed taxonomy.  The central parent node represents the overarching AI Robotic System, from which three primary branches emerge: Attack Surface, Ethical and Legal Concerns, and Human-Robot Interaction. Attack surfaces is further into two families: physical attack and digital attack. Physical attacks encompass attacks on the perception layer (input sensors). Digital attacks include attacks on navigation and planning, attacks on actuators, and attacks on 
  {AI} models (training and inference attacks).} 
  \label{fig:taxonomy} 
\end{figure*}

{We begin the survey by categorizing the AI-Robotic systems into three 
{dimensions}, including \textit{attack surfaces, ethical and legal concerns}, and \textit{human-robot interaction security}, as illustrated in our taxonomy (see Fig. \ref{fig:taxonomy}). The first category explores attack surfaces and their impact on the architectural components of robotic systems. The second category, on the other hand, discusses the ethical and legal concerns surrounding the use of robotics systems in various domains such as healthcare, the military, autonomous vehicles, manufacturing, and agriculture. Finally, we explore the security landscape of human-robot interaction and recommend defensive strategies to safeguard privacy concerns.}

The attack surfaces are divided into three distinct domains: \textit{physical attack, digital attack}, and \textit{other relevant attacks}. Physical attack encompasses attacks targeting the perception layer (attack in input sensors, see Section \ref{section_sensor_attack}) 
within the hybrid robotic architecture. 
{Digital attack, on the other hand,} pertains to attacks directed at the navigation and planning, as well as the control layer. The interconnectedness between physical and digital attacks becomes evident, with physical attacks having a direct impact on digital systems. For example, if input sensors such as cameras, LiDAR, or RGBD sensors are tampered with, they could produce poisoned data. When these compromised data (see Section \ref{section_training_attack}) are used by digital systems like object recognition algorithms, it may lead to misclassifications, thereby affecting navigation and planning (see Section \ref{section_navigation_and_planning}). 
{Furthermore, it can also cause mistimed actuation commands within the control systems} (see Section \ref{section_actuator_attack}). On the other hand, the manipulation of physical objects within the environment can introduce a new set of challenges, potentially leading to deception, malfunctioning, or erratic behaviors in robotic systems. A compelling illustration of this is the work by Eykholt et al. \cite{eykholt2018robust}, who showcased physical attacks using adversarial stickers to deliberately deceive an autonomous vehicle's object recognition system. This illustrates how vulnerabilities in physical components (both environment and sensors) can cascade into the digital realm, causing major disruptions and interruptions in robotic system operation. Furthermore, AI-Robotics systems can also suffer from inherent challenges of AI/ML models itself such as model training attacks (see Section \ref{section_training_attack}) and model inference attacks (see Section \ref{section_inference_attack}). 

The third category within the attack surfaces is other relevant attacks. These include backdoors, malware, ransomware, as well as network and communication-based attacks on robotics systems. We discuss these attacks in greater detail in Section \ref{other_relevant_attacks}.

Following this, our focus extends to the ethical and legal concerns (see Section \ref{section_ethical_and_legal}) surrounding the use of these systems. We explore the ethical dilemmas arising from robotics technology in various domains such as healthcare, autonomous vehicles, assistive robots, military, and manufacturing.
{The focus is to showcase the importance of ethical considerations and their influence in the development, deployment, and use of AI-Robotics systems in the aforementioned domains. For instance, healthcare robotics systems such as surgical robots must be developed that align with medical ethics, ensuring patient safety and privacy. Another important issue to consider is how to program AI agents in complex ethical scenarios. Examples of this include the trolley problem and the self-sacrifice dilemma as depicted in Fig }\ref{fig:trolley_problem} in relation to Autonomous Vehicles (AVs). 
  {Aside from these concerns, the overt reliance on autonomous robotic agents for decision-making processes can have adverse effects on human cognitive thinking and reasoning abilities, ultimately leading to a reduction in human autonomy \cite{datteri2009ethical}.} From legal point of view, we explore two regulations that ensures privacy protection to personal data. The first one is California Consumer Privacy Act (CCPA\cite{CCPA} in case of the United States (US) and General Data Protection Regulations (GDPR)\cite{gdpr} in case of Europe (EU). Both CCPA and GDPR provide similar privacy protections to consumers.

{Furthermore,} we expand our survey to encompass the realm of Human-Robot Interaction (HRI) 
{security} (see Section \ref{section_HRI}), delving into the intricacies of 
{physical, cognitive and social interactions between people and robots.} 
{We survey current approaches that are aimed at safeguarding interactions between humans and AI-Robotic systems.} 


The survey of attack surfaces and their defensive solutions based on the taxonomy is available in Section \ref{section_attack_surfaces} while the survey of ethical and legal concerns surrounding AI-Robotic systems is in Section \ref{section_ethical_and_legal}, and human-robot interaction in Section \ref{section_HRI}. 
\section{Attack Surfaces}
\label{section_attack_surfaces}

    In the context of robotic systems, an attack surface is a set of potentially vulnerable points on the boundary of a system, a system element, or an environment, through which an adversary attempts to compromise or manipulate the robot's functionalities \cite{ross2019protecting}. Attack surfaces on robotic platforms can be broadly divided into two categories such as \textit{physical attack} and \textit{digital attack} \cite{hollerer2021cobot} 
    {as illustrated in Fig \ref{fig:taxonomy}.} 


\subsection{Physical Attack}
\label{section_sensor_attack}
Physical attacks involve the act of physically manipulating hardware or components with the aim of either disrupting its normal functioning or obtaining unauthorized access. For example, tampering, destruction, removal, theft, or manipulation of embedded sensors disrupt the perception of robotic systems. Humanoid and social robots that interact with human counterparts are susceptible to these attacks, as evidenced by the authors in \cite{oruma2022systematic}. On the other hand, hardware-level trojans presented in \cite{bhunia2014hardware} can manipulate physical components such as integrated circuits and cause robotic systems to malfunction abruptly. 



{Over the next few paragraphs, we will investigate an array of attack vectors encompassing techniques like jamming and spoofing in relation to input sensors. Our analysis will also extend to how these vectors influence the perception layer within hybrid AI-Robotics systems architecture.}



The perception layer is the first component of AI-Robotics architecture. It serves as a vital pillar in AI-Robotics systems, utilizing an array of {proprioceptive and exteroceptive sensors. Proprioceptive sensors such as Inertial Measurement Unit (IMU), Global Positioning System (GPS), encoders, etc. are responsible for measuring values internal to the robotic system and exteroceptive sensors such as cameras, Light Detection And Ranging (LiDAR), radar, and others, are responsible for collecting information regarding the robot's environment. These sensors' information is utilized to comprehend and operate through the environments.} These sensors constitute physical devices, and any attempts to compromise their integrity fall within the realm of physical attacks, as elaborated in our taxonomy. They play a pivotal role in enabling AI-Robotics systems to create a comprehensive world model, facilitating collision avoidance, precise path and trajectory planning, and accurate location mapping. However, the perception layer becomes a natural target for adversaries, who aim to tamper with or compromise the quality of sensor data \cite{deng2021deep}. Manipulation of sensor data affects the integrity and reliability of the perception process, thereby degrading the performance of the navigation and planning layers, and ultimately influencing the behaviors in the control layer. 
{A summary of the attack surface, affected sensors, attack summary, and proposed defensive strategies can be found in Table \ref{table:Attack_sensors_}}.

\begin{figure}[htb]
  \centering
  \includegraphics[width=1.0\linewidth]{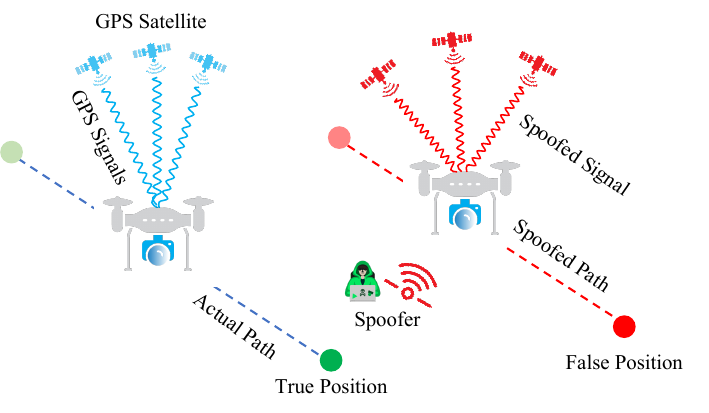}
  \caption{An illustration of GPS spoofing attack on UAVs, in which an adversary redirects the UAV to a false location by transmitting deceptive GPS signals.}
  \label{fig:sensor_attack}
\end{figure}

\begin{table*}[!htbp]
\caption{{Overview of attacks on input sensors along with defensive strategies.}}
\centering
{\renewcommand{\arraystretch}{1.20}%
\newcolumntype{s}{>{\columncolor{lightgray!20!}}p{2cm}}
\begin{tabular}{s|p{2.0cm}|p{5.5cm}|p{4.5cm}|p{1.5cm}}
    \hline
    \rowcolor{cyan!20!}
    Attack Surface & Medium & Attack Summary & Defensive Strategy & Research \\  \hline
        \multirow{4}{*}{Jamming} & \multirow{4}{*}{\shortstack[l]{Camera, GPS, \\LiDAR, \\ Ultrasonic \\ sensors}} & 
        \begin{minipage}[t]{\linewidth}      
         \begin{itemize}[leftmargin=*]
             \item Use of intense light to jam camera sensors.
             \item Using bright light source (blinding light) to render  LiDAR renders the sensor ineffective in perceiving objects. 
          \end{itemize}       \vspace{1mm}      
         \end{minipage}
         & 
         \begin{minipage}[t]{\linewidth}      
         \begin{itemize}[leftmargin=*]
             \item Incorporating a near-infrared-cut filter and photochromic lenses to selectively filter certain types of light. 
          \end{itemize}       \vspace{1mm}      
         \end{minipage} 
         & \cite{yan2016can, shin2017illusion, lim2018autonomous, kar2014detection}\\ \hline

       \multirow{5}{*}{Spoofing} & \multirow{5}{*}{GPS, LiDAR} & 
        \begin{minipage}[t]{\linewidth}      
         \begin{itemize}[leftmargin=*]
             \item GPS spoofing attack designed to manipulate navigation system.
             \item Adversary may spoof LiDAR sensors using techniques such as laser projection\cite{cao2019adversarial, shin2017illusion, jin2022pla}, shape manipulation\cite{cao2021invisible}, and object manipulation \cite{zhu2021can}. 
          \end{itemize}       \vspace{1mm}      
         \end{minipage}
         & 
         \begin{minipage}[t]{\linewidth}      
         \begin{itemize}[leftmargin=*]
             \item Filtering \cite{kune2013ghost}, Randomization \cite{shin2017illusion}, and Fusion \cite{xu2018analyzing}. 
             \item Anomaly detection sytems \cite{Rivera2019AutoEncodingRS}. 
             \item Spoofing detection systems for radars-based spoofing attacks \cite{Kapoor2018DetectingAM}.
          \end{itemize}       \vspace{1mm}      
         \end{minipage} 
         & \cite{zeng2018all}, \cite{cao2019adversarial, shin2017illusion, jin2022pla, cao2021invisible}\\ \hline

         \multirow{4}{*}{Manipulation} & \multirow{4}{*}{IMUs} & 
        \begin{minipage}[t]{\linewidth}      
         \begin{itemize}[leftmargin=*]
             \item Manipulate the gyroscope, accelerometer, and magnetometer sensors.
             \item Disrupt the accurate measurement of linear and angular velocity and cause errors in control system. 
          \end{itemize}       \vspace{1mm}      
         \end{minipage}
         & 
         \begin{minipage}[t]{\linewidth}      
         \begin{itemize}[leftmargin=*]
             \item Inertial sensors safeguarding.
             \item Sensor fusion techniques. 
          \end{itemize}       \vspace{1mm}      
         \end{minipage} 
         & \cite{ji2021poltergeist}\\  \hline

         \multirow{3}{*}{\shortstack[l]{Software\\ vulnerabilities}} & OpenCV, TensorFlow Image, Open3D & 
        \begin{minipage}[t]{\linewidth}      
         \begin{itemize}[leftmargin=*]
             \item Adversary may exploit software vulnerabilities within programs responsible for processing sensor data.
             
          \end{itemize}       \vspace{1mm}      
         \end{minipage}
         & 
         \begin{minipage}[t]{\linewidth}      
         \begin{itemize}[leftmargin=*]
             \item Data verification and validation.
             \item Data Sanitization. 
             \item Patching and updating for known vulnerabilities. 
          \end{itemize}       \vspace{1mm}      
         \end{minipage} 
         & \cite{bradski2000opencv, abadi2016tensorflow, zhou2018open3d}\\ \hline

\end{tabular}}

\label{table:Attack_sensors_}
\vspace{-5mm}

\end{table*}
In the realm of adversarial attacks, adversaries can engage in various tactics such as {jamming, spoofing, manipulating, or software vulnerability exploitation for proprioceptive and exteroceptive input sensors, causing disruptions, impairments, and deceptions that hinder the accurate perception of the environment by these sensors.}

\subsubsection{Jamming Attacks}
In \cite{yan2016can}, the authors discuss the use of intense light to jam camera sensors, leading to a significant deterioration in the quality of captured images and subsequently affecting object recognition. Another study in \cite{shin2017illusion} focuses on blinding attacks targeted at LiDAR sensors, where a bright light source with the same wavelength as the LiDAR renders the sensor ineffective in perceiving objects from the direction of the light source. Similarly, in \cite{lim2018autonomous} researchers simulate jamming attacks on ultrasonic sensors, while the vulnerability of GPS signals to jamming devices is investigated in \cite{kar2014detection}.
These attacks could also lead to other types of attacks, such as a Denial-of-Service (DoS) attack wherein the perception sensors become unavailable for authorized users as a result of malicious jamming.

\subsubsection{Spoofing Attacks}
Recent studies suggest adversaries can also spoof LiDAR sensors including three techniques such as laser projection techniques \cite{cao2019adversarial, shin2017illusion, jin2022pla}, shape manipulation strategies \cite{cao2021invisible}, and object manipulation approaches \cite{zhu2021can}. An example of GPS spoofing attack is presented in \cite{zeng2018all} wherein a GPS spoofing attack is meticulously designed to manipulate navigation systems.
For instance, in Figure \ref{fig:sensor_attack}, a GPS spoofing attack on a UAV is depicted. In this scenario, the attacker manipulates the path trajectory and alters the true position of the UAV by transmitting false or spoofed GPS signals.

\subsubsection{Manipulation Attacks}
In addition, some studies have brought to light the vulnerability of IMUs employed in robotic systems to targeted attacks. These attacks aim to manipulate the gyroscope, accelerometer, and magnetometer sensors \cite{nashimoto2018sensor}. Such attacks aim to disrupt the accurate measurement of linear and angular velocity, consequently causing the robotic systems to lose control and in some cases deceive object detection tasks \cite{ji2021poltergeist}. 

\subsubsection{Software Vulnerability Attacks}
In addition to the aforementioned attacks, robotic systems are susceptible to software vulnerabilities within programs responsible for processing sensor data, such as OpenCV \cite{bradski2000opencv}, TensorFlow Image \cite{abadi2016tensorflow}, and Open3D \cite{zhou2018open3d}. Despite their widespread use, these frameworks have certain limitations and vulnerabilities, as they often rely on assumptions regarding the structure and format of sensor data. As a result, if these assumptions are not met, these frameworks may fail \cite{goodfellow2014explaining}. For example, OpenCV has been discovered to be vulnerable to a heap-based buffer overflow (CVE-2018-5268) \cite{cve1}, which can be triggered by parsing a malicious image during the loading process. Similarly, vulnerabilities can arise from the lack of, or incomplete, or invalid validation of input parameters (CVE-2021-37650) \cite{cve2} in the TensorFlow library, assuming that all records in the examined dataset are strings. 

To mitigate blinding attacks on cameras, a defense strategy involves incorporating a near-infrared-cut filter during daytime operations \cite{petit2015remote}. {Also,} the utilizing photochromic lenses to selectively filter certain types of light {can enhance} the overall image quality. In a separate study \cite{kar2014detection}, the authors propose a method for detecting GPS jamming incidents. Similarly, countermeasures against spoofing attacks encompass techniques such as filtering \cite{kune2013ghost}, randomization \cite{shin2017illusion}, and fusion \cite{xu2018analyzing} to enhance the robustness and reliability of the sensors. Additionally, the advancement of machine learning techniques has led to the development of spoofing detection methods, as demonstrated in \cite{Rivera2019AutoEncodingRS}. This research paper introduces an anomaly detection approach that employs an auto-encoder architecture specifically designed for LiDAR sensors to identify sensor spoofing attacks. Furthermore, {a} proposal in \cite{Kapoor2018DetectingAM} outlines a strategy to detect and counter radar-based spoofing attacks by implementing a spatio-temporal challenge-response technique. This method verifies analog domain physical signals, effectively preventing sensor spoofing incidents. Besides, verification and sanity checking of input sensor data should be conducted to ensure input data are free from adversarial inputs.

{Next, we delve into digital attacks, where we explore attack patterns on the navigation and planning layers and control layers (actuators) of a hybrid AI-Robotic architecture. Furthermore, we also discuss the inherent challenges of ML models.}

\subsection{Digital Attack}
Digital attack pertains to software-related attacks that aim to exploit the software vulnerability within the robot’s operating system, algorithms, firmware, communication channels, or other software component (for example AI-based agents) to manipulate navigation \& planning (see Section \ref{section_navigation_and_planning}) and control layer (see Section \ref{section_actuator_attack}) of a robotic systems. The literature also shows AI-specific attacks such as training attacks (see Section \ref{section_training_attack}) and inference attacks (see Section \ref{section_inference_attack}) that impact robotic systems.

In the subsequent subsections, we present a comprehensive analysis of attack patterns and their associated security strategies within the diverse components of AI-Robotics systems.

\subsubsection{Navigation and Planning Attacks with Security Strategies}

\label{section_navigation_and_planning}

Attacks on the navigation and planning layer of robotic system architecture present a complex and concerning challenge in the realm of AI, automation, and cybersecurity. Contrary to the perception-oriented attacks through robotic input sensors, where the objective is to tamper with {perception} sensors. 
{Attacks on navigation and planning relate to manipulating the AI-Robotics system’s navigation systems by leveraging the poisoned data generated from compromised sensors.} With the rapid integration of autonomous robots into daily human lives, from self-driving cars to industrial automation, 
{from military applications to health care,} ensuring the security and integrity of their navigation and planning systems becomes paramount. 
{Table \ref{table:navigation_planning_summary} provides an overview of various forms of attacks on the navigation and planning layers of the hybrid AI-Robotics system architecture, along with defense strategies.}

Navigational attacks can manifest in various forms, ranging from simple and localized disruptions to controlled, sophisticated, widespread manipulations. One of the most common forms of attack involves sensor spoofing or jamming \cite{xu2022sok, ryu2015review}, where adversaries intentionally deceive the robot's sensors, such as GPS, camera, LiDAR, etc, to provide false or distorted information. Doing so can mislead the robot's perception of its environment, leading to potential collisions, misrouting, or compromised decision-making. Furthermore, navigational systems could be vulnerable to data manipulation attacks \cite{bhardwaj2019cyber}. These attacks involve altering the mapping or localization data robots use, causing incorrect interpretations of the environment. For example, in an urban setting, this could result in a robot delivering packages to the wrong address or navigating hazardous routes, endangering pedestrians and other vehicles.

\begin{table*}[!htbp]
\caption{{Research summary on adversarial attacks on robotic navigation and planning systems with proposed defense strategies.}}
\centering
{\renewcommand{\arraystretch}{1.20}%
\newcolumntype{s}{>{\columncolor{lightgray!20!}}p{2cm}}
\begin{tabular}{s|p{3cm}|p{4.5cm}|p{4.5cm}|p{2cm}}
    \hline
    \rowcolor{cyan!20!}
    Attack Surface & Medium & Attack Summary & Defense Strategies & Research \\  \hline

        
        & Communication Sensor (GPS, Radio, etc,) & 
        \begin{minipage}[t]{\linewidth}      
         \begin{itemize}[leftmargin=*]
             \item Transmit spoofed positional data.
             \item Block communication channel. 
          \end{itemize}       \vspace{1mm}      
         \end{minipage}
         & 
         \begin{minipage}[t]{\linewidth}      
         \begin{itemize}[leftmargin=*]
             \item Need minimal control input for scope limitation.
             \item Need secured recovery algorithm.
          \end{itemize}       \vspace{1mm}      
         \end{minipage} 
         & \cite{bianchin2019secure, yang2020enhanced, dey2018security, he2019govern, he2018friendly, sathaye2022experimental, zeng2018all, shen2020drift, xu2022sok}\\ \cline{2-5}

        & Visual Sensor (Camera, LiDAR, etc,) &      
        \begin{minipage}[t]{\linewidth}      
         \begin{itemize}[leftmargin=*]
             \item Falsify environment information for incorrect perception.
             \item Manipulate operational environment.
          \end{itemize}       \vspace{1mm}      
         \end{minipage}
         &
         \begin{minipage}[t]{\linewidth}      
         \begin{itemize}[leftmargin=*]
             \item Incorporate adversarial training.
             \item Need to use advanced robust AI algorithm.
          \end{itemize}       \vspace{1mm}      
         \end{minipage}
         & \cite{li2021fooling, tang2021fooling, petit2015remote, shin2017illusion, jin2023pla, sun2020towards, cao2021invisible, zhu2021can, ji2021poltergeist, xu2022sok} \\ \cline{2-5}
        
        \multirow{-7}{*}{Navigation} & Environmental Force Sensor (IMU, etc,) &
        \begin{minipage}[t]{\linewidth}      
         \begin{itemize}[leftmargin=*]
             \item Alters operational environment.
          \end{itemize}       \vspace{1mm}      
         \end{minipage}
         &
         \begin{minipage}[t]{\linewidth}      
         \begin{itemize}[leftmargin=*]
             \item Reduce single-sensor decision dependency.
          \end{itemize}       \vspace{1mm}      
         \end{minipage} 
         & \cite{son2015rocking, trippel2017walnut, nashimoto2018sensor, tu2018injected, xu2022sok} \\  \hline

        
        & Black-box DRL models and planners (Image Segmentation, 3-D point cloud, etc,) &
        \begin{minipage}[t]{\linewidth}      
         \begin{itemize}[leftmargin=*]
             \item Exploit and optimize the cost function for controlled degradation.
          \end{itemize}       \vspace{1mm}      
        \end{minipage}  
        & 
        \begin{minipage}[t]{\linewidth}      
         \begin{itemize}[leftmargin=*]
             \item Through careful tuning and environment-specific modifications are needed.
             \item Development on non-iterative planning schemes.
          \end{itemize}       \vspace{1mm}      
        \end{minipage}  
        & \cite{vemprala2021adversarial, rempe2022generating, li2020survey, shen2020drift, bhandarkar2022adversarial, hickling2022robust, peng2012intelligent, wang2021can} \\ \cline{2-5}

        \multirow{-2}{*}{Planning} & Coordination algorithms (Robust \cite{zhou2018resilient, saldana2017resilient} , Adaptive
and reactive \cite{ramachandran2019resilience, song2020care}) &
        \begin{minipage}[t]{\linewidth}      
         \begin{itemize}[leftmargin=*]
             \item Break communication for coordinated systems and prevent recovery.
             \item Exploit risk-reward trade-off algorithms. 
          \end{itemize}       \vspace{1mm}      
         \end{minipage} 
         & 
         \begin{minipage}[t]{\linewidth}      
         \begin{itemize}[leftmargin=*]
             \item Utilize GNN algorithms for decentralized information propagation.
             \item Research on risk/uncertainty-aware search algorithms.
          \end{itemize}       \vspace{1mm}      
         \end{minipage} 
         & \cite{zhou2021multi}, \cite{park2018robust, yang2017algorithm, zhou2018approximation, tolstaya2020learning, li2020graph, he2018friendly, sathaye2022experimental, shen2020drift, li2021chronos} \\ \hline 

\end{tabular}}

\label{table:navigation_planning_summary}


\end{table*}

To prevent sensor spoofing by the adversary and enable secure and optimized navigation, Bianchin et al. \cite{bianchin2019secure} developed an algorithm to detect spoofed sensor readings and falsified control inputs intended to maximize the traversing trajectory. The authors consider an undetectable attack over Global Navigation Satellite System (GNSS) sensor and a Radio Signal Strength Indicator (RSSI) sensor. The detection algorithm works on the principle of nominal control input. 
In another work by Yang et al. \cite{yang2020enhanced}, the authors introduced timing-based adversarial strategies against a DRL-based navigation system by jamming in physical noise patterns on the selected time frames. The attack imposes adversarial timing based on three principle scenarios: pulsed zero-out attack (noise through off-the-shelf hardware), Gaussian average on sensor fusion (noisy sensor fusion system developed by Gaussian filter), and adversarial noise patterns (fast gradient sign method to generate adversarial pattern against trained Deep Q-network (DQN)). Through the attack, both value and policy-based DRL algorithms are easily manipulated by a black-box adversarial attacking agent. As a preventive measure, the authors suggested adapting adversarial training to train DQN and Asynchronous Advantage Actor-Critic (A3C) models by noisy states.

On the other hand, 
{attacks on planning phase} are equally disruptive. 
{ It targets} 
algorithms responsible for determining a robot's actions and decisions based on sensory information. Adversaries could introduce subtle changes to the robot's decision-making process, causing it to make sub-optimal choices or even hazardous actions \cite{wang2022resilient, peng2012intelligent}. For example, an autonomous drone might be tricked into flying into a restricted area or performing unauthorized surveillance. Such actions can be in the form of a targeted signal or sensor distortion or in the form of data manipulation. In the work of Wang et al. \cite{wang2023resilient}, the authors proposed jamming attacks using 
{Reinforcement Learning} (RL) to influence the 
{Unmanned Aerial Vehicle} (UAV) performance and then developed anti-jamming strategies using 
double 
{DQN} to mitigate the attack 
in UAV trajectory planning. 

{Similarly,} Bhandarkar et al. \cite{bhandarkar2022adversarial} proposed 
DRL techniques to 
{launch} Sybil attacks and transmit spoofed beacon signals to disrupt the path planning logic. 
Hickling et al. \cite{hickling2022robust}, 
 {on the other hand,} proposed to utilize the explainability of DL methods. In 
 {their} approach, the planning agent is trained with a Deep Deterministic Policy Gradient (DDPG) with Prioritized Experience Replay (PER) DRL scheme, that utilizes Artificial Potential Field (APF) to improve training time and obstacle avoidance performance. 
 {Along the same lines}, Peng et al. \cite{peng2012intelligent} proposed an Online Path Planning (OPP) framework to continuously update the environmental information for the planner in adversarial settings. For intelligently selecting the best path from the OPP output, the Bayesian network and fuzzy logic are used to quantify the bias of each optimization objective.

Additionally, SLAM is the problem of planning and controlling the motion of a robot to build the most accurate and complete model of the surrounding environment \cite{placed2023survey}. Active SLAM problems fall under Partially Observable Markov Decision Processes (POMDPs), where both action and observation uncertainties affect navigation and planning. In several cases, adversaries exploit such uncertainty to disrupt the autonomous navigation and planning system. In the work by Wang et al. \cite{wang2021can}, the authors introduced an infrared light-based attack that can alter environment perception results and introduce SLAM errors to an autonomous vehicle. For defensive measures, the authors developed a software-based detection module. Similarly, Li et al. \cite{li2021chronos} demonstrated an attack that exploits timing-sensitive multimodal data from different sensors to cause destabilization of the SLAM perception model in cyber-physical systems. Researchers have emphasized various recommendations to prevent and mitigate beyond-line-of-sight and active spatial perception errors from getting exploited \cite{placed2023survey} and require active research to mature and achieve real-world impact.


{In the next section, we explore attacks on the control layer of hybrid AI-Robotics systems architecture with a focus on attacks on actuators and recommend corresponding security strategies.}

 \begin{figure}[t]
  \centering
  \includegraphics[width=\linewidth]{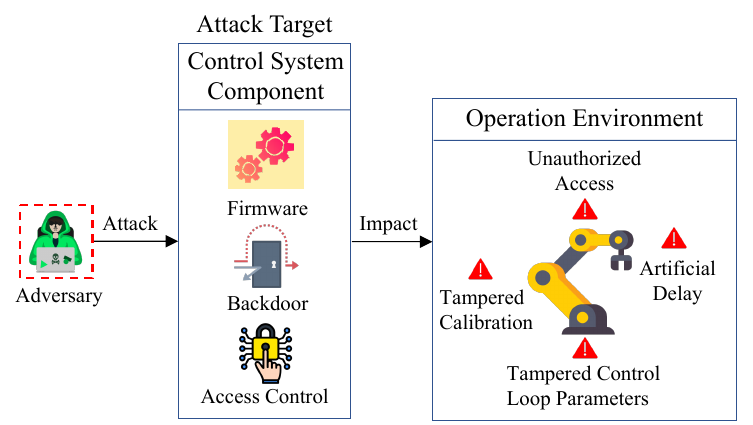}
  \caption{An overview of attack surfaces in control systems and their impact on actuators within the operating environment. Icons from \cite{flaticon}.}
  \label{fig:actuator_attack}
\end{figure}

\subsubsection{Actuator Attacks with Security Strategies}

\label{section_actuator_attack}


In the robotic systems architecture,  robots act on the environment by leveraging its joint manipulators (in the case of robot arms or humanoids), or wheels (in the case of ground mobile robots) or propellers (in the case of aerial and marine robots). To act, these systems need actuators. An Actuator is a mechanical device that utilizes various sources of energy, such as hydraulics, external signals, pneumatics, and electricity, to convert them into motion that can be manipulated according to specified control parameters \cite{herath2022foundations}. Electric motors, hydraulic actuators, and pneumatic actuators are the most common types of actuators used in robotics. 
{As highlighted in the Section \ref{survey_approach} and depicted in the taxonomy Figure \ref{fig:taxonomy}, attacks on actuators are classified as digital attacks.}

Potential attack surfaces for robot actuators include actuator hardware and firmware, communication protocols, and network components such as those used by tele-operated robots.
{Fig \ref{fig:actuator_attack} illustrates the attack scenario wherein the adversary targets the firmware, manipulates the access control mechanism, or employs backdoor techniques to influence the behavior of robotic systems within their operating environment.} Table \ref{table:attack_actuators_} summarizes various types of attacks on actuators along with defensive strategies. To gain access to the robot control systems, attackers can install hardware-level backdoors or kill switches to bypass control protection mechanisms \cite{wang2012software}. For example, Sasi \cite{sasi2015maldrone} developed a backdoor program that allows attackers to remotely control 
UAV robotic systems. 
Another area that is vulnerable to attack is actuator firmware, largely due to insecure communication (for example, using the public internet to update a robot’s firmware). Several studies \cite{vilches2018introducing, kirschgens2018robot, pogliani2019security} indicate that attackers can exploit communication vulnerabilities of public networks to gain full control of robotic systems.

Authentication (for example Sybil attack \cite{douceur2002sybil}) and access control are often overlooked in robotic systems. Researchers indicate that some robotic application interfaces are designed without the need for a login portal, allowing anyone to access them remotely \cite{lacava2021cybsersecurity}, while others that are equipped with access control systems are also known to suffer from vulnerabilities. These include communication packets containing credentials in clear text or weakly encrypted, systems with default or hard-coded credentials that can be found in publicly accessible user manuals, or in some cases, a lack of access rights configuration allowing adversaries with low privileges to control robotic systems \cite{7958582}.

Recent scientific research has shown the potential vulnerability of the control systems and actuators of teleoperated surgical robots to malicious attacks. In their study, Bonaci et al. \cite{Bonaci2015ToMA}, demonstrated the ability to exert unauthorized control over various surgical functions. Furthermore, various research has revealed targeted attacks that can bypass the robot's input sensors, including the surgeon's input, and directly impact the actuators \cite{Alemzadeh2016TargetedAO}. These attacks result in behaviors such as the involuntary halting of the system or the abrupt movement of robotic arms. Notably, similar attacks on actuators have been observed in other robotic applications, such as mobile robotics \cite{Guo2017ExploitingPD, Gao2020StateEA}, robotic Cyber-Physical Systems (CPS) \cite{Giraldo2020DARIADA, Huang2018ReliableCP}, and aircraft systems \cite{He2021HMMbasedAA}.




Security strategies for robot actuators involve several measures to ensure the safety and integrity of the robot's operations. One crucial aspect is the sanitization of control signals that are transmitted to the actuators, which helps protect against adversarial inputs. To achieve this, concepts like dynamic filtering \cite{escudero2022enforcing} can be applied to filter out anomalous control signals. Moreover, an access control mechanism can also be implemented to ensure only verified programs send control signals to the actuator driver. Another approach is to develop anomaly detection systems to monitor anomalies during robot manipulation tasks. For example, the authors in \cite{park2016multimodal}, presented a method for multimodal anomaly detection during robot manipulation. In a separate study, the authors in \cite{guo2018roboads} developed RoboADS, a robot anomaly detection system that is capable of detecting both sensor and actuator misbehaviors. The alerts generated by these detection models can then be analyzed by robot operators to conduct further investigations if necessary. Furthermore, rule-based actions can be implemented in the actuator code to prohibit the robot from executing undesired actions.

{Additionally, attacks on robotic actuators possess significant security concerns regarding the kinematics and dynamics factors due to their ability to interfere with the environment. From the kinematic and dynamics standpoint, compromised actuators can generate motions or forces intending to harm their sensors, machinery, and peripherals. For example, in an industrial setting, actuator attacks causing kinematics disruption can destroy the manufacturing pipeline, instruments, and many more. In the work of Zhang et al}.~\cite{zhang2023kinematic},{ the authors demonstrated mitigation strategies against actuator attacks for kinematic disruption in networked industrial systems that can cause a decrease in manufacturing quality. Again, Moaed et al.}~\cite{abd2019simulated} {demonstrated attacks on actuator dynamics and restoration process through attacks on prosthetic arm actuators, which can disrupt human electromyogram control signals and develop frustration and distrust in the human mind. Furthermore, the casualty from a kinematic or dynamic impact is not limited to the production pipeline or human mind— a carefully triggered kinematics and dynamics attack in the heavy manufacturing (nuclear power plant, chemical gas plant, steel plant, and other) industry can incur catastrophic casualties such as loss of lives. Therefore, the security of actuators requires significant attention when integrated with AI technologies for operations. }

{Moving forward, we examine the inherent challenges within AI models and their potential impacts on AI-Robotics systems.}

\begin{table*}[!htbp]
\caption{{Overview of attack patterns in actuators and associated defensive strategies.}}
\centering
{\renewcommand{\arraystretch}{1.20}%
\newcolumntype{s}{>{\columncolor{lightgray!20!}}p{2cm}}
\begin{tabular}{s|p{2.0cm}|p{5.5cm}|p{4.5cm}|p{1.5cm}}
    \hline
    \rowcolor{cyan!20!}
    Attack Surface & Medium & Attack Summary & Defensive Strategy & Research \\  \hline
    \multirow{5}{*}{Hardware} & \multirow{5}{*}{\shortstack[l]{Backdoors and \\ kill switch}} & 

    \begin{minipage}[t]{\linewidth}      
         \begin{itemize}[leftmargin=*]
             \item Installation of hardware-level backdoors or kill switches to bypass control protection mechanisms.
             \item Backdoor programs to remotely control robotic systems, e.g., UAVs. 
          \end{itemize}       \vspace{1mm}      
         \end{minipage} 
    & 
     \begin{minipage}[t]{\linewidth}      
         \begin{itemize}[leftmargin=*]
             \item Secure design and development. 
             \item Utilizing formal verification methods to ensure the robotic system's software and hardware meet specified security requirements.
          \end{itemize}       \vspace{1mm}      
         \end{minipage} 
    & \cite{wang2012software, sasi2015maldrone}\\ \hline

    \multirow{5}{*}{Firmware} & \multirow{5}{*}{\shortstack[l]{Insecure \\ network \\ communication}} &  

     \begin{minipage}[t]{\linewidth}      
         \begin{itemize}[leftmargin=*]
             \item Attackers can exploit communication vulnerabilities of public networks to install malware and obtain full control of robotic system. 
          \end{itemize}       \vspace{1mm}      
         \end{minipage} 
    & 
     \begin{minipage}[t]{\linewidth}      
         \begin{itemize}[leftmargin=*]
              \item Employing secure communication protocols and encryption mechanisms between robotic components.
             \item Update firmware and patch software using secured connection. 
          \end{itemize}       \vspace{1mm}      
         \end{minipage} 
     & \cite{vilches2018introducing, kirschgens2018robot, pogliani2019security}\\ \hline

    \multirow{4}{*}{Access Controls} & Access controls mechanism and authentication systems &  

        \begin{minipage}[t]{\linewidth}      
         \begin{itemize}[leftmargin=*]
              \item Lack of proper access control mechanism allowing anyone to access and control remotely.
          \end{itemize}       \vspace{1mm}      
         \end{minipage} 
 
    & 
        \begin{minipage}[t]{\linewidth}      
         \begin{itemize}[leftmargin=*]
             \item Inclusion of robust access control to ensure only verified programs send control signals to the actuator driver.
          \end{itemize}       \vspace{1mm}      
         \end{minipage}

    & \cite{lacava2021cybsersecurity, 7958582}\\ \hline

    \multirow{4}{*}{Others} & Manipulation of surgical functions (surgical robots) &  
    \begin{minipage}[t]{\linewidth}      
         \begin{itemize}[leftmargin=*]
             \item Targeted attacks can bypass the robot's input sensors, including the surgeon's input, and directly impact functioning of actuators.
          \end{itemize}       \vspace{1mm}      
         \end{minipage} 
    & 
    \begin{minipage}[t]{\linewidth}      
         \begin{itemize}[leftmargin=*]
             \item Sanitization of control signals. 
             \item Dynamic filtering \cite{escudero2022enforcing} can be applied to filter out anomalous control signals. 
          \end{itemize}       \vspace{1mm}      
         \end{minipage} 
    & \cite{Bonaci2015ToMA, Alemzadeh2016TargetedAO}\\ \hline
\end{tabular}}

\label{table:attack_actuators_}
\vspace{-5mm}
\end{table*}

\subsubsection{Attack on AI Models} 
Modern robots contain multiple software components such as object recognition \cite{browatzki2012}, voice commands \cite{alonso2011}, and facial recognition \cite{cruz2008}, that are usually built using 
{AI}-based pipelines. At a high level, 
AI pipeline entails data collection, preprocessing, model selection, training, validation, and deployment. These 
{AI} pipelines can be used to solve previously intractable problems in robotics using adaptable and data-driven methodologies. However, this same adaptability can also lead to exploitation by attackers \cite{barreno2010} because of the inherent complexity and the many stages involved in training, validating, and deploying an 
{AI} system. 
{AI} components that involve deep learning (i.e. neural networks) can be especially vulnerable due to their ``black box'' nature, which means that even when the security is compromised, it may be difficult to pinpoint exactly where the problem occurred and how to address it \cite{castelvecchi2016}. The nature of this attack is classified as a digital attack, as delineated in our taxonomy in Section \ref{section_taxonomy} and depicted in Fig \ref{fig:taxonomy}.

{In the following subsections, we explore training and inference attacks on 
{AI} models.}  

\label{section_attack_ML_model}

\paragraph{Training Attacks and Security Strategies}
\label{section_training_attack}

\begin{figure*}[t]
  \centering
  \includegraphics[width=\linewidth]{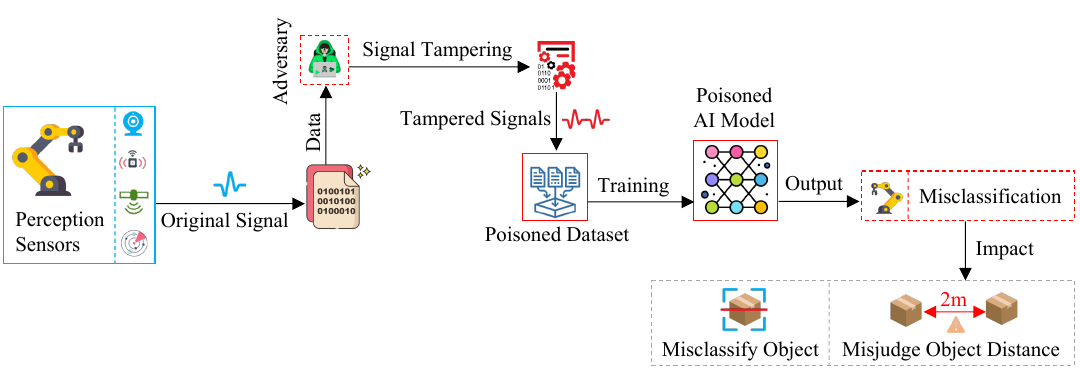}
  \caption{An example of training attack on 
  {AI} pipeline that causes misclassification of objects and misjudgment of object distance.}
  \label{fig:training_attack}
\end{figure*}

Training attacks involve the deliberate alteration, tainting, and modification of the training dataset. 
{Such an attack intends} to manipulate the 
training process and compromise the performance and behavior of 
{AI} models.  There are several forms of security attacks on the training process or training data on 
{AI} agents of AI-Robotic systems. One form of attack is known as \textit{model poisoning}, where the training data for 
{AI} models is tainted through label corruption. 
Such attacks might lead 
{AI} model to function unexpectedly or is unable to recognize the current input data. Figure \ref{fig:training_attack} presents a high-level overview of a training attack. The adversary tampers sensory data obtained from perception sensors such as cameras, IMUs, LiDAR, etc., to create a poisoned dataset, which is then utilized to train an 
{AI} model. As a result, the trained 
{AI} model is compromised and may exhibit misclassifications, such as inaccurately identifying objects or misjudging distances between objects. There are several well-known model poisoning cases such as the Microsoft Tay chatbot incident \cite{tays2019} and the VirusTotal poisoning incident \cite{virustotal2021}. The Tay chatbot incident involved an experimental conversational chatbot released by Microsoft on Twitter, which is a popular social media platform. Tay is designed to work by learning from real-world conversations and improving its language capabilities over time. Unfortunately, malicious internet users exploited this property by coordinating to send messages to it involving racist, misogynistic, and anti-semitic language. As a result, Tay started to repeat these abusive and offensive messages and the experiment had to be suspended. Another example is the VirusTotal poisoning incident, where mutant samples uploaded to the training platform were able to fool an ML model into incorrectly classifying inputs. VirusTotal is a popular platform for analyzing suspected malware files and automatically sharing them with the security community. However, attackers were able to compromise the integrity of this platform by uploading fake samples such that the platform would classify even benign files as malicious. In essence, these model poisoning cases show that an 
{AI} model can be compromised if the source of its training data is not validated. Robotic systems that utilize publicly-sourced training data can be vulnerable to these model poisoning attacks. In addition, outsourced training can 
{potentially} introduce the risk 
{of malicious} backdoors 
\cite{Liu2018FinePruningDA}.

One strategy to prevent model poisoning attacks 
{is} either 
{preventing} a model from being poisoned or repairing a network after it is poisoned \cite{Goldblum2020DatasetSF}. 
{Another strategy involves} identifying poisoned data as outliers in the input space \cite{charikar2017, severi2021}, employing certain input preprocessing techniques \cite{liu2017}, or by trimming out irregular samples when computing the loss function \cite{Jagielski2018ManipulatingML}. 
{In the event of poisoned model, it may be} 
possible to detect the planted artifacts using generative adversarial networks \cite{Zhu2020GangSweepSO} and repair them using pruning and fine-tuning \cite{Liu2018FinePruningDA}. Pruning works by removing neurons that are non-essential or dormant for clean inputs, thus reducing the risk that the network will produce unwanted behavior for bad inputs. On the other hand, fine-tuning works by taking a network trained with untrusted sources and then further training the network with trusted inputs. A combination of the two, fine-pruning \cite{Liu2018FinePruningDA}, can even defeat more sophisticated pruning-aware attacks.

Building upon the exploration of training attacks and associated security strategies, in the next section, we delve into inference attacks and their corresponding security strategies.

\paragraph{Inference Attacks and Security Strategies}
\label{section_inference_attack}

\begin{figure}[htb]
  \centering
  \includegraphics[width=\linewidth]{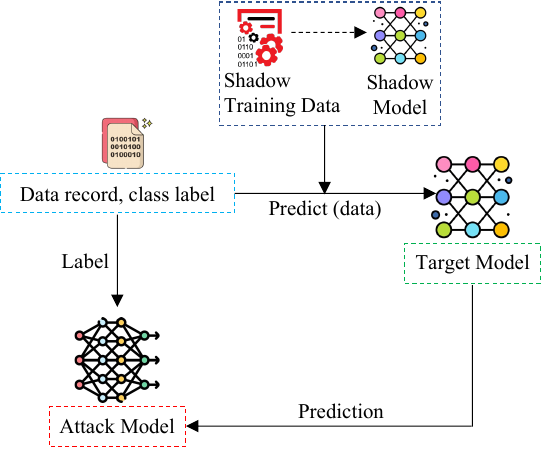}
  \caption{An overview of a membership inference attack where the attacker queries the target model utilizing the shadow training set and shadow model to obtain a prediction vector containing probabilities for each class \cite{shokri2017membership}. }
  \label{fig:inference_attack}
\end{figure}

A model inference attack is a type of privacy attack wherein an adversary attempts to reverse-engineer or infer sensitive information about the training data or individual data points used to train an 
{AI} model by querying the model and analyzing its outputs \cite{ shokri2017membership}. 
Construction of model inference attack usually involves generating shadow models (a replica of the target model that mimics its behavior as closely as possible) using shadow training \cite{shokri2017membership}. An example of a model inference attack is depicted in Figure \ref{fig:inference_attack}. Inference attacks on AI-Robotic systems may include model inversion, model evasion, and model alteration among others. 

\begin{itemize}
    \item \textit{Model Inversion} attacks also known as attribute inference attack \cite{fredrikson2015model} occur when an adversary strategically queries the inference API to extract potentially private information embedded within the training data \cite{Kaya2021WhenDD}.
Research has shown that certain data augmentation techniques can mitigate overfitting and indirectly reduce the risk of membership inference attacks \cite{shokri2017membership} but such methods may incur an accuracy penalty \cite{Kaya2021WhenDD}. 
\item \textit{Model Evasion} tactics are used by attackers to prevent the 
{AI} model from computing the correct output. In the image domain, adversarial techniques can take the form of simple transformation of the input (cropping, shearing, translation), adding white noise in the background, or constructing adversarial examples by carefully perturbing the input to achieve desired output \cite{machado2021}. In the natural language domain, adversarial techniques can take the form of a Happy String, wherein the benign input is tacked onto malicious query points, or by using bad characters, words, or sentences that can still appear semantically sound from a human perspective \cite{Chen2020BadNLBA}. Certain model evasion strategies can be especially stealthy since they employ optimization techniques to find the smallest perturbations that can still cause a change in the model output \cite{Kashyap2021ReLUSynSS}.
\item \textit{Model Alteration} occurs when attackers are able to gain access to the network weights and alter them without the knowledge of the user. These attacks are essentially deployment-stage backdoor attacks where the attacker can cause the network to produce incorrect output for a specific input while producing correct outputs for most other inputs \cite{Bai2021TargetedAA, Qi2021SubnetRD}. Such attacks can also cause neural networks to become black-box trojans. Since the behavior of a neural network can significantly change while appearing benign from the outside if only minimal model parameters are changed \cite{Pan2020BlackboxTO, Bai2021TargetedAA}. These 
{attacks} can take the form of network structure modification \cite{Pan2020BlackboxTO}, flipping weight bits\cite{Bai2021TargetedAA}, or subnet replacement \cite{Qi2021SubnetRD}.

\end{itemize}

Defense strategies to preserve the privacy of users' data (in case of membership inference attacks), especially images for computer vision problems, is to apply a reconstruction and perturbation network. This method preserves image features while obfuscating the individual's face \cite{Wen2021IdentityDPDP, Maximov2020CIAGANCI}. Differential Privacy (DP) is another technique that can mitigate privacy issue by computing a theoretical bound on the effect that any particular input has on the output. Employing such strategy may provide guarantees in certain situations that information will not be leaked \cite{Bguelin2020AnalyzingIL}. Another technique to prevent data leakage is learning with selective forgetting \cite{Shibata2021LearningWS}, where a class-specific synthetic signal is used to forget arbitrary classes without using the original data. In a similar vein, defense against model evasion techniques that utilize adversarial inputs is to formally check the security properties of a deep neural network with Satisfiability Modulo Theory or symbolic interval analysis \cite{Wang2018FormalSA}. In the case of evasion attacks, two forms of defense can be implemented. The first type focuses on improving an AI model's robustness by introducing methods that make the model's weights and layers more difficult to steal or reverse-engineer \cite{orekondy2019prediction}. This can be accomplished via online learning and weight readjustment or by introducing bounded perturbations \cite{orekondy2019prediction}. The second form employs a hybrid approach. This involves integrating AI algorithms to deal with unknown quantities with tried and tested methods from legacy systems, which are designed to tackle known robotics problems \cite{ashkenazy2019cylance}. This approach provides an extra degree of verification to ensure comprehensive protection.

\subsection{Other Relevant Attacks}
\label{other_relevant_attacks}

As highlighted in Section \ref{section_taxonomy}, AI-Robotics systems can also suffer from other common cybersecurity attacks. Based on our taxonomic Figure \ref{fig:taxonomy}, some of these attacks include backdoors, malware, ransomware, and attacks on communication networks. In the following paragraphs, we explain these attacks in detail.

A \textit{backdoor attack} is a stealthy technique of circumventing normal authentication methods to gain unauthorized access to a system. Malicious actors can insert a backdoor in robotic systems to remotely control them. In some cases, threat actors such as malicious manufacturers may leave, backdoor, on purpose, in robotic systems to track and monitor the activities of the robot and its operator without the owner’s knowledge ~\cite{yaacoub2022robotics}. 

{\textit{Malware}, on the other hand, can target robotic systems such as surgical robots. Chung et al.\cite{chung2019smart}, for example, exploited ROS vulnerabilities and developed smart self-learning malware capable of tracking the movements of the robot's arms and triggering the attack payload during a surgical procedure.} Similarly, \textit{ransomware attacks} in the context of industrial robots attempt to lock them to extort ransom from the manufacturer. In an effort to augment robot control system security, the authors in \cite{mayoral2020industrial} presented a customized industrial robot ransomware, which showcased the possibility of industrial robots being susceptible to ransomware attacks. 

Another potential candidate of attack 
is the robotic \textit{communication network}. Adversaries may target vulnerability of existing communication protocols which might lead to jamming, de-authentication, traffic analysis, eavesdropping, false data injection, denial of service, replay, or man-in-the-middle attacks. Jamming and de-authentication attacks are quite common in small UAVs \cite{krishna2017review}. Traffic analysis and eavesdropping are passive attacks used on the unencrypted channels of UAVs to extract the meaningful and sensitive flight and motion data \cite{mekdad2023survey} whereas false data injection and replay attacks involve insertion of incorrect routing information into the caches of UAVs \cite{mekdad2023survey,wu2007survey}.



\section{Ethical and Legal Concerns}
\label{section_ethical_and_legal}




Apart from the attack surfaces discussed in preceding section, there are also ethical and legal concerns surrounding robots. 
{In this section, we will explore these aspects in greater detail.} 

\subsection{Ethical Concerns}

One of the popular topics that have an ethical dimension in robotics domain is \textit{robot ethics} or  \textit{roboethics} - a term first coined by  Gianmarco Verrugio \cite{veruggio2005birth}. It is a combination of ethics (\textit{what’s right and wrong}) and robotics (robots and technology). It focuses on comprehending and overseeing the ethical outcomes and aftermaths arising from robotics technology, especially with regard to intelligent and autonomous robots, within our societal context \cite{robot_ethics}. Roboethics encompasses a diverse range of application areas where ethical considerations play a pivotal role in shaping the development, deployment, and use of robots and autonomous systems. 

In following paragraphs, we explain ethical concerns surrounding use of AI-Robotics systems across various domains including healthcare, autonomous vehicles, assistive robots, military, and manufacturing as delineated in Section \ref{section_taxonomy} and depicted in Fig. \ref{fig:taxonomy}. 

In the realm of healthcare, medical robotic technologies are increasingly utilized for surgeries \cite{leonard2014smart}, rehabilitation \cite{ai2021machine}, and caregiving. Ethical questions arise concerning patient privacy, cost of implementation, and informed consent for robotic procedures among others. 
{Consider, for instance, the substantial expense associated with robotic surgery. This naturally prompts us to ask}\cite{tzafestas2016roboethics}: 
{ ``Given that there is marginal benefit from using robots,
is it ethical to impose financial burden on patients or the medical system?''} 
{Other ethical quandaries include patients' autonomy and raise questions such as how to ensure patients fully understand and consent to robotic procedures? Do they have the right to opt for or against robotic assistance in surgery? Similarly, there are questions about data privacy and security. Questions like, What safeguards are in place to protect patient data collected and transmitted by medical robots? How do we ensure privacy and prevent data breaches?}

\textit{Medical roboethics} guides the development of these systems to ensure patient safety, privacy, and equitable access. 
{Furthermore, the design and development of medical robots must comply with privacy rules, including the Health Insurance Portability and Accountability Act (HIPAA)} \cite{act1996health} and the European Union’s General Data Protection Regulation (EU GDPR) \cite{regulation2016regulation}. 
{These regulations guide engineers and developers in ensuring the privacy and security of user data (see Subsection \ref{privacy_concern_around_robots} for privacy concerns around human robot interaction). By adhering to these policies, developers can mitigate the risk of data breaches, unauthorized access, and misuse of sensitive medical information.}


\begin{figure}[htb]
  \centering
  \includegraphics[width=.70\linewidth]{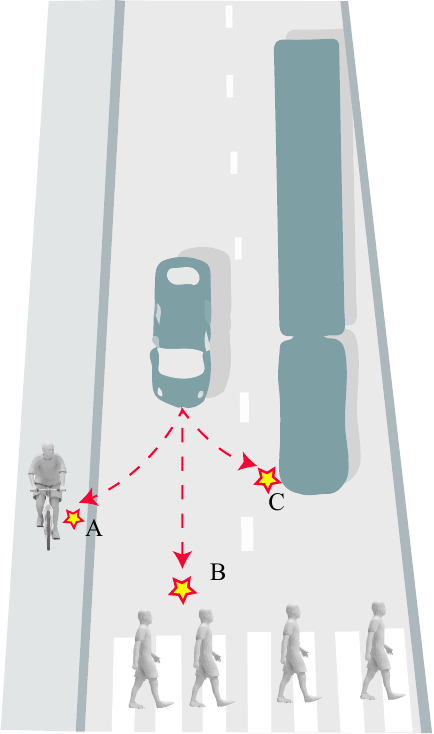}
  \caption{A graphical depiction of the ethical dilemma for the AVs in an \textit{inevitable crash}. Option (A): crash with the cyclist on the right; Option (B): continue straight and crash with two pedestrians; Option (C): self-sacrifice by crashing with the truck on the left. Adapted from \cite{wang2020ethical}. 
  }
  \label{fig:trolley_problem} 
\end{figure} 
Another critical area of robotic application is Autonomous Vehicles (AV). There are two fundamental ethical concerns surrounding AVs, including the trolley problem and the self-sacrifice dilemma \cite{wang2020ethical}. 
{A visual illustration of these dilemma is presented in Fig} \ref{fig:trolley_problem}. 
The first one refers to a self-driving car deciding whether to continue on its present path, potentially hitting and harming multiple pedestrians, or to swerve and divert its path, potentially killing a single individual \cite{goodall2016can, goodman2017european}.  Then, in such a scenario, what is the correct decision? It raises questions about how these decisions should be programmed by AI agents, who bears the ethical responsibility, or how societal values and cultural norms should influence the behavior of these vehicles. The self-sacrifice dilemma, on the other hand, is a variant of the trolley problem that addresses the ethical question of whether an autonomous vehicle should prioritize the safety of its occupants (passengers) over the safety of pedestrians or other individuals outside the vehicle. In this scenario, the self-driving car is confronted with an unavoidable collision and must determine whether to take actions that would minimize harm to its passengers at the expense of greater harm to pedestrians or vice versa \cite{lin2013ethics}. Then again, what is the correct decision? Should the AI agents be programmed in a way that saves more people by sacrificing the driver? How can these decisions be made transparent? This issue underscores the need to consider integrating societal values, legal frameworks, and cultural norms in designing AI agents that can intelligently adapt their behavior in such ethically complex situations.

In a similar vein, assistive robots (for example rehabilitation robots, robotized wheelchairs, walkers, etc.) designed to assist people with disabilities, or the elderly are becoming increasingly prevalent. These robots can improve autonomy and quality of life, but ethical quandaries arise surrounding questions of dependency, risk, cost, benefits, and the potential for social isolation induced by growing reliance on machines for companionship and care. 
{For example, to what extent should individuals using assistive robots maintain control over the actions and decisions of the robot?, who is liable in case of accidents or errors made by assistive robots?, and how can we maintain a balance between the benefits of companionship and the potential risks of emotional dependence \cite{riek2014code} on machines?.}

The use of sophisticated and automated robotic weapons by the military in conflict has profound ethical concerns. Particularly, lethal autonomous weapons raise concerns about moral accountability for machine actions, and the potential dehumanization of combat. 
{For instance, how can we ensure meaningful human involvement in critical decisions, especially in life-and-death situations?, how can we program AI agents that distinguish combatants and non-combatants? how can we ensure that targeting decisions made by AI agents aligns with humanitarian principles and laws? This is a critical issue because, during the fog of war, it is difficult enough for humans to effectively distinguish whether a target is legitimate. So, how can we ensure AI-Robotics systems perform better ethical decision-making than human soldiers?}\cite{mcdaniel2008robot}.

Additionally, in the industries like manufacturing and agriculture, ethical considerations surrounding human-robot collaboration, job displacement, and responsible automation are central to the roboethics discourse.

\subsection{Legal Concerns}


Laws in robotics are influenced by fictional writers such as Isaac Asimov. His three laws of robotics \cite{asimov1941three}, which first appeared in a story called \textit{``Runaround''}, include the following:

\begin{enumerate}
    \item A robot may not harm a human being, or, through inaction, allow a human being to come to harm.
\item A robot must obey the orders given to it by human beings, except where such orders would
conflict with the First Law.
\item A robot must protect its own existence, as long as such protection does not conflict with the First
or Second Law.
\end{enumerate}
{While the laws straightforwardly establish robots as subordinate to humans, their implementation can be challenging in diverse real-world scenarios. Nevertheless, robot engineers can view them as guidelines for programming AI agents within robotic systems.} Legal and regulatory concerns stem from the application of legal and moral theories pertaining to liability and responsibility \cite{asaro2016liability}. Consider the utilization of surgical robots in hostile military environments for procedures like ``remote robotic-assisted telesurgery’’ or ``autonomous robotic surgery’’.  In the event of an unintended event mid-surgery, where harm is inflicted upon human counterparts due to malfunctioning or attack on robot actuators, this may result in legal consequences. This prompts a discussion regarding which party should be held liable?, such as the system manufacturer, software developer, the autonomous agents involved (this is difficult because as they are not legal persons), or the human operators overseeing the operation of these devices \cite{o2019legal}.



{Another legal dimension within the AI-Robotics systems in security of personal data. To safeguard the personal information various policy framework has been proposed both in the United States (US) known as 2020 California Consumer Privacy Act (CCPA) \cite{CCPA}) and 2016 General Data Protection Regulation (GDPR \cite{gdpr}) in the Europe (EU). 

Before CCPA, in the US, the Executive Office of the President, National Science and
Technology Council, Committee on Technology issued a report, ‘‘Preparing for the Future of Artificial Intelligence’’ (‘‘AI Report’’), in October 2016} \cite{obama}. 
{Given the importance of AI in the future, the report recommended ‘‘the Department of Transportation should work with industry and researchers on ways to increase sharing of data for safety, research, and other purposes.’’  Prior to this, the same administration also presented a discussion draft on “Consumer Privacy Bill of Rights Act”. It proposed seven principles including transparency; individual control; respect for context; focused collection and responsible use; security; access and accuracy; and accountability, however, the act was not enacted} \cite{ishii2019comparative}.
To alleviate the concerns of use of personal data by large technology companies CCPA was brought to legislation in 2018. The regulation defines \textit{personal data and information}, enforce requirements for \textit{company data security, allow data portability} and \textit{deletion rights}, and \textit{impose fines} on organizations that violate them. This law went to effect in 2020, after undergoing several amendments. The CCPA is based on a consumer protection framework and applies more narrowly to enterprises that meet specified revenue or number of California residents' personal information collection requirements \cite{wong2023privacy}.

{Similarly, in the EU, GDPR law was signed in April 2016 and implemented in May 2016. The new GDPR mandates explicit informed consent for data use. Through informed consent, the users are given the right to opt in or out from any potential processing of their data through the ``right of explanation.'' GDPR also empowers users to track or remove their data through the ``right to be forgotten.'' To counter the problem of information leakage, the authors in \cite{Graves2021AmnesiacML} presented two efficient approaches, such as unlearning and amnesiac unlearning. Their approach is successful in removing personal data. In a similar vein, Villaronga et al. \cite{Villaronga2018HumansFM} examined the issue of AI memory in the context of the ''right to be forgotten'' and explored the inadequacies of current privacy laws for tackling the complexities and challenges of AI technology. The paper analyzed legal implications for AI as well as the conflict between privacy and transparency under EU privacy law. In a separate study, the authors in \cite{Golatkar2020EternalSO} proposed an approach that cleanses the weights in a deep neural network to forget information selectively.}

In the subsequent section, we delve into challenges and risks that arise in the context of human-robot interaction.



\section{Human-Robot Interaction (HRI) Security Studies}
\label{section_HRI}

Human-Robot Interaction (HRI) is an emerging field that examines the physical, cognitive, and social interactions between people and robots. 
While HRI is an important component of robotic autonomy, studies have also identified several key vulnerabilities that can compromise the integrity, privacy, and safety of these interactions. For example, unauthorized access to a collaborative robot (co-bot) in a manufacturing setting could involve potential manipulation of the robot’s behavior. Similarly, privacy is another major concern, as the collection and storage of personal and sensitive information during interactions with social robots used in healthcare settings can result in privacy breaches or unauthorized use of data. {It is important to note that interactions can occur between people and collaborative multi-robot systems (MRS) including robot swarms.}

{In accordance with the taxonomy introduced in Section \ref{section_taxonomy}, we conduct a survey of the challenges and concerns in the subsequent subsections.} 
{We also} recommend 
countermeasures such as 
{Privacy Enhancing Technologies (PET)}  that 
bolsters the integrity, privacy and confidentiality of data exchanged between humans and robots.


\subsection{Challenges and Concerns}

\subsubsection{Privacy Concerns Around Robots}
\label{privacy_concern_around_robots}
User privacy needs to be considered and protected when designing robotic assistive systems. Privacy concerns might influence human behavior when robots monitor them. In their work \cite{kelly2012privacy}, the authors analyzed how older adults perceive privacy and their attitudes to engage in privacy-enhancing behavior under three natural settings -- a camera, a stationary robot, and a mobile robot. Authors found that more participants engaged in more privacy-enhancing behavior such as self-censoring of speech or covering up the camera when being watched by a camera compared to embodied robots. Their analysis implies that participants are less cautious about their privacy around robots due to their unfamiliarity with them.

Robots endanger people's privacy by directly monitoring them and increasing access to their information. A study in \cite{reinhardt2021still}, examined 1090 German-speaking older adults verifying the factors that influence their privacy concerns and preferences. The paper observed a significant correlation between the people with practical robotic experience and the interest. The participants interested in human robots showed higher comfort levels around robots than the users who selected machine robots. Most participants did not express interest in sharing their health-related information with anonymous health services. Results indicated that participants prefer not to transmit their location and position to others unless there is an emergency. Furthermore, their level of comfort varied depends on the data collection modalities. 

\subsubsection{Social Acceptance To Robot Threats}

People exposed to robots that are capable of rejecting human commands could result in a negative attitude towards robots. In the context of their experimental investigation \cite{ zlotowski2017can}, the authors investigated social acceptance of perceived robot autonomy in case of threats to human identity and threats to jobs. Using the Mechanical Turk crowdsourcing platform, they surveyed 176 US participants showing videos of autonomous robots that could disregard human commands and non-autonomous robots that could only obey human commands. Participants were assessed on their perception of robotic threats to jobs, resources, safety, and human identity. They also evaluated participants' negative attitudes towards robots and support for the robotic research. The survey concluded that the participants viewed autonomous robots as more realistic threats than those that were not autonomous. Generally, humans are more hesitant to trust 
especially those they perceive to be deceptive \cite{ doi:10.1177/1064804611415045, wagner2011acting}. However, within their study \cite{kontogiorgos2020embodiment} found that robotic failures in their experiments did not negatively affect user attitude when using a human-like robot embodiment. Furthermore, as outlined in their research endeavor \cite{mirnig2017err}, researchers confirm that error-prone robots are more believable.

\subsubsection{Internet-Connected Devices}
Storing and sharing personal information of parents and children is a potential threat in internet-connected 
{robotic} toys. 
Within the scope of empirically evaluating the security and privacy perspective about parent attitudes and child privacy, the authors in \cite{10.1145/3025453.3025735}, made recommendations for toy designers related to the security and privacy of the toys. They focused on identifying parents' privacy expectations and concerns,  children’s mental models of their privacy, and the ethical perspective of parental controls. They conducted semi-structured interviews with thirty parents and children, following a demonstration and interaction session with two different toys. Authors suggested that 
{robotic} toy designers must be transparent and communicate with parents that the toy is recording and delete recordings after a fixed time from the app and the cloud servers. Such studies aid designers in devising internet-connected games and toys without violating privacy rights.


{On the other hand, smart homes, more specifically, Robot-integrated Smart Homes (RiSH) \cite{do2018rish}, are on the rise. These smart homes integrate a home service robot, a home sensor network, a body sensor network, a mobile device, cloud servers, and remote caregivers. RiSH can be used for research in assistive technologies for elderly care. The interconnectivity of these systems through the Internet of Things (IoT), however, exposes them to various cybersecurity attacks and poses a number of security concerns.} In their research \cite{205156}, the authors collected responses from semi-structured interviews with fifteen smart home residents to identify the gap between risks identified by security researchers and the user's security concerns and their mitigation methods. 
{The findings showcased smart home users' lack of security concerns, ranging from not feeling personally targeted or trusting potentially adversarial actors to believing existing mitigation strategies to be sufficient.} 

\subsubsection{Trust}

Psychologists define trust as a mental state in humans \cite{atoyan2006trust}. In relation to HRI, there are various definitions of trust; for example, Lee et al. \cite{lee2004trust}, define it as ``the attitude that an agent will help achieve an individual's goals in a situation characterized by uncertainty and vulnerability'' whereas Hancock et al. \cite{hancock2011can}, defines it as ``the reliance by an agent that actions prejudicial to their well-being will not be undertaken by influential others''. The key takeaway from these concepts is \textit{whether a robot's activities and behaviors align with human interests} \cite{khavas2020modeling}. 

In their work \cite{hancock2011can}, the authors identified three categories of factors that affect trust in HRI: robot-related factors, task and environmental factors, and human-related factors. The first category pertains to the quality of operation performed by the robot from the perspective of the human operator and includes subtasks \cite{khavas2020modeling} such as transparency and feedback in decision-making \cite{salem2015would}, situational awareness \cite{boyce2015effects}, autonomy \cite{lazanyi2017dispositional}, fault tolerance \cite{desai2012effects}, and engagement \cite{robinette2013building} to facilitate naturalistic interaction. In contrast, task and environment-related factors, as identified in various other studies, such as the nature of the task \cite{tay2014stereotypes}, risk and human safety \cite{robinette2016overtrust}, and the task site \cite{lazanyi2017dispositional}, influence trust between robots and humans. The third category, on the other hand, is concerned with human-based factors, such as the general perception about robots \cite{obaid2016stop}, expectations of human operators towards automation \cite{yang2017evaluating}, past experience with robots \cite{walters2011long}, and operators' understanding of robotic systems \cite{ososky2013building}.

\subsection{Privacy Enhancing Technologies}



\begin{figure}[htb]
  \centering
  \includegraphics[width=\linewidth]{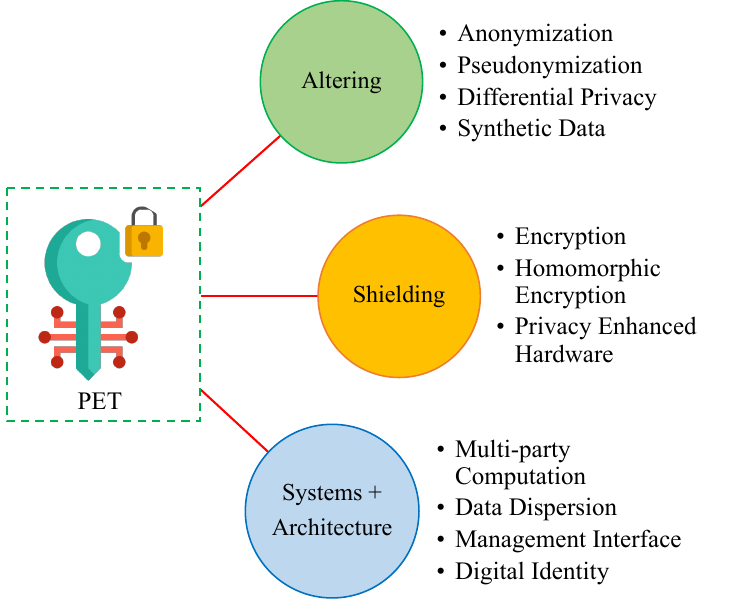}
  \caption{A graphical illustration of Privacy Enhancing Technologies (PET). Adapted from \cite{pet}.}
  \label{fig:pet_technology} 
\end{figure}

{In the context of AI-Robotics, Privacy Enhancing Technologies (PET) refer to the use of various technological solutions and strategies to protect individuals' privacy when interacting with robotic systems. Their purpose is to protect user identities by offering anonymity, pseudonymity, unobservability, and unlinkability of both users and  data subjects} \cite{fischer2001security , pfitzmann2005anonymity}. 
{A report by the Federal Reserve Bank of San Francisco} \cite{pet} 
{categorizes PETs in three categories: \textit{altering data, shielding data, and systems + architecture}. A graphical depiction of these categories and subcategories is presented in Fig. \ref{fig:pet_technology}. The first category of PETs are tools that alter data itself. Examples of these include anonymization, pseudonymization, differential privacy, and synthetic data. The second category focuses on hiding and shielding the data. This usually involves the use of cryptographic techniques. Some examples include encryption, homomorphic encryption, and privacy-enhanced hardware. The final category of PETs are new systems and processes for data activities. The report suggested that some of these systems also enable greater transparency and oversight across data activities, including collection, processing, transfer, use, and storage.}

{Authors in} \cite{kaaniche2020privacy} 
{surveyed PETs for solving personalization paradox. Their approach included much broader PETs as compared to the report by Federal Reserve Bank of San Francisco. Some of these technologies include anti-tracing (fingerprinting), data obfuscation (perturbation, private information retrieval), privacy preservation certification (attribute-based signatures, sanitizable signatures), self-destructing data systems (vanish, neuralyzer), statistical disclosure control (anonymizing DBs), and secure communication (client-service encryption, end-to-end encryption).}

{PET such as} \textit{access control mechanisms} are popular in achieving privacy and data protection in natural settings. Furthermore, the right choice of sensors that adapt to differing privacy expectations is essential in reducing privacy vulnerabilities \cite{eick2020enhancing}. To address this, the authors in \cite{horstmann2020towards} proposed an app-based user interface to enable users to be in control of their data. The interface is intended to enable transparency and accuracy in data processing by communicating with the software running on the robot. This transparent interface displayed the robot's visual, auditory, position, and communication information to help users identify the possible privacy violations caused by social robots. However, the human feedback evaluation revealed that participants had somewhat mixed attitudes towards this self-learning social robot. Moreover, the work in \cite{butler2015privacy} recommends balancing privacy with utility in robot systems. The qualitative and quantitative evaluations of privacy preferences of end-users in their study have shown concerns about both privacy and physical harm.

In relation to trust in HRI, to enhance trust in robotic applications, trust models must be developed for a specific form of human-robot interaction or a specific type of robotic agent \cite{khavas2020modeling}. For example, a trust model for evacuation robots must be different from a trust model for surgical robots because the expected outcomes of these domain-specific robotic systems are distinct.




\section{Future Research and Discussion}
\label{section_research_direction}

Previous sections discuss the security risks and challenges faced by AI-Robotics systems. 
{More, specifically, we explored three dimensions of AI-Robotics systems including, \textit{attack surfaces, ethical and legal concerns,} and \textit{security of HRI}. As these AI-Robotics systems become more pervasive, their security becomes critical. We believe that the holistic understanding of abovementioned dimensions  will effectively strengthen the overall security posture of these systems. 
Security of input sensors and sensory data from these sensors (e.g., training data, models) becomes essential to ensuring the integrity of these systems. The data complexity only increases as you consider that an AI-Robotics system could be comprised of many robots (i.e., a system of systems). We see future research focusing on ensuring trustworthy AI-Robotics systems that are capable of withstand various attack scenarios, adhere to the ethical and legal principles of the domain in which they are deployed, and ensure the interactions with these systems remain private.}

\begin{figure}[htb]
  \centering
  \includegraphics[width=1\linewidth]{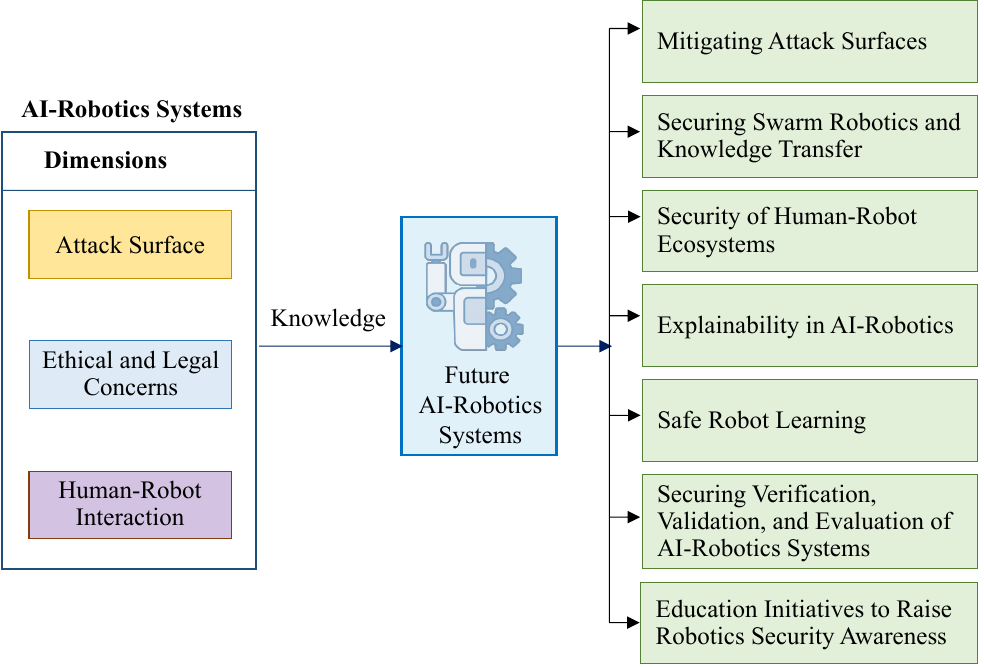}
  \caption{A visual representation outlining the future research direction aimed at improving the security of AI-Robotics systems.}
  \label{fig:future_direction} 
\end{figure}

{In the following subsections, we outline seven essential domains that encompass various research areas critical for enhancing the security of AI-Robotics systems. As depicted in Fig \ref{fig:future_direction} these include research on attack surfaces mitigation, securing swarm robotics and knowledge transfer, security of human-robot ecosystems, explainability in ai-robotics, securing verification, validation, and evaluation of ai-robotics systems, andeducation initiatives to raise robotics security awareness.}

\subsection{Mitigating Attack Surfaces}

{In Section \ref{section_attack_surfaces}, we explored various attack surfaces on AI-Robotics systems. The first component that an adversary may target is the embedded sensors that are responsible to percept the environment. Adversaries can either physically damage sensors or they can manipulate sensors by jamming or spoofing the sensor signals. The future robotic systems should be designed with adversaries’ capability in mind. For example, to secure the sensors from a physical attack such as vandalism defenders can apply perimeter security with robust access controls. Similarly, in case of other attack patterns such as jamming and spoofing the defenders may integrate lightweight spoofing and jamming detection algorithms \cite{Rivera2019AutoEncodingRS, Kapoor2018DetectingAM}. Another important aspect that future AI-Robotics systems designers and engineers must consider is to have some mechanism for verification and sanity checking of input sensors data. Such process would thwart adversaries attempt to supply adversarial input to AI agents. }

{To secure the actuators from attacks, the research is leaning towards development of anomaly detection systems. These systems are capable of monitoring the anomalies during robot manipulation task \cite{park2016multimodal}. Such methods can also be extended to filter out anomalies in sensor data \cite{ guo2018roboads}. Furthermore, control signal sanitization mechanism must be integrated in robotics systems of the future to defend against adversarial input to actuators. }

{Future security strategies to secure the AI agents used by AI-Robtics systems during training phase may include certain preprocessing techniques \cite{liu2017} such as identifying poisoned data as an outlier in the input space \cite{charikar2017, severi2021}. Furthermore, the robotic systems can engineers leverage generative models such as Generative Adversarial Networks to detect the planted artifacts in poisoned model \cite{Zhu2020GangSweepSO}. In case of protecting privacy during inference attack various security strategies can be implemented for future robotic application. Concepts like differential privacy, selective forgetting \cite{ Shibata2021LearningWS}, data obfuscation \cite{Wen2021IdentityDPDP, Maximov2020CIAGANCI} can be considered. }

\subsection{Securing Swarm Robotics and Knowledge Transfer}
As noted before, robots are becoming more integrated in various aspects of human life, and securing human-robot interactions is going to be vital moving forward. Similarly, securing the interactions between robots in multi-robot systems is another major challenge. Advances in swarm robotics \cite{dorigo2020reflections} have led to large-scale swarm applications in which teams of robots work together to achieve a common goal. {The need to share} information across agents has led to new architectures \cite{venkata2023kt} including making use of shared-memory structures called blackboards. Swarms may find themselves inoperative if a constituent is targeted by an adversarial attack. {Decentralized security solutions, including secured federated learning \cite{shan2021towards} and cybersecurity knowledge graphs \cite{mitra2021combating}, are becoming more prominent in order to protect swarms from agent-specific targeted attacks.}


\subsection{Security of Human-Robot Ecosystems}
In human-robot ecosystems where all agents work independently, a major challenge is preventing conflicts between robotic agents and humans utilizing shared resources \cite{chakraborti2016planning}. \textit{How are plans prioritized? How is information exchanged? Can negotiations occur?}

In scenarios where AI-Robotics systems assist humans 
{(i.e., HRI applications)}, \textit{trusting} inferences made by these robotic systems is another challenge \cite{kok2020trust}. Sensors in robots could become faulty or biased over time and produce inaccurate measurements. \textit{How does a robot re-calibrate a faulty sensor in an autonomous setting? How do you fuse information in a multi-robot system that may or may not have a faulty sensor (or a faulty robot)?} These are important questions that must be addressed in order to \textit{trust} these systems in scenarios with high variability and where human life is at stake.

It is important to note that in human-robot ecosystems, agents can be both physical and digital. In particular, agents can be a digital representation of a physical system (i.e., digital twin\cite{fuller2020digital}). Digital twins are being used to improve human-robot collaboration in manufacturing \cite{bilberg2019digital}. Improving the robotic system through human experience and knowledge is also an important area of research. An emerging concept is the use of a human digital twin (HDT) in an elaborate system-of-systems where human decision making is captured by data-driven approaches \cite{wang2022human}, \cite{neupane2023twinexplainer}. 

\subsection{Explainability in AI-Robotics}
Explainable AI (XAI) provides a mechanism for securing 
{HRI applications}. 
Roque et al. \cite{roque2022explainable} describe and implement an explainable AI solution that can be used by humans to calibrate their trust in their systems. While most current XAI solutions are geared towards domain experts \cite{neupane2022explainable}, other approaches \cite{das_xai_robot_failure} target non-experts when explaining robot failures. In order to provide truly comprehensive XAI solutions in AI-Robotics, end-users with no (or very little) AI/ML experience must be considered. In fact, the majority of end-users will not be experts in many robotic applications.

Another emerging research topic is neurorobotics. Since the beginning, AI and robotics have always drawn inspiration from biology. Today, researchers \cite{chen2020neurorobots} are leveraging neuroethology, the combination of neuroscience and ethology, to study the effect of neural network perturbations in hopes of explaining a robot's behavior.

\subsection{Safe Robot Learning}
The increased use of data-driven approaches in safety-critical applications has raised the question of safety (and safe guarantees) for such learning algorithms \cite{fulton_safe_rl}. Brunke et al. \cite{brunke2022safe} define \textit{safe learning control} as the problem of safe decision-making under uncertainties using machine learning. Reinforcement learning (i.e., data-driven) approaches are highly adaptable but do not provide safety guarantees. Conversely, control theory (i.e., model-driven) approaches provide guarantees within known operation conditions but are less adaptable. Current research \cite{ berkenkamp2017safe} is investigating the merger of control theory and reinforcement learning to provide generalizable and safe solutions within certain boundaries.

Many of these solutions use physics-based environments during model training. In certain scenarios, these environments will also have to be protected to avoid corruption and unintended disclosure of information (e.g., performance bounds, physical constraints) for any derived model.

\subsection{Securing Verification, Validation, and Evaluation of AI-Robotics Systems}
Ensuring safe, secure, and trustworthy AI is an active area of research \cite{thiebes2021trustworthy}. This naturally extends to AI-Robotics systems. \textit{How can these systems be verified and validated in large-dimensional and dynamic environments? How do you ensure that these systems are robust to situations different from those encountered during the training/design phase (i.e., new/unseen data)? How do you capture all the decisions made by an AI-Robotics system for traceability?} The future of AI-Robotics will look at mitigating security risks and data protection.

\subsection{Education Initiatives to Raise Robotics Security Awareness}


{Raising robotics security awareness and training is crucial in our increasingly interconnected world, where robots and autonomous systems are becoming more prevalent.} 

Hands-on hacking exercises on various forms of cyber-attacks on Robotic systems are an effective tool for security education to raise awareness about various forms of cyberattacks to Robotic systems stakeholders. For example, within the scope of their scholarly investigation \cite{votipka2021hacked}, 
{the authors reviewed the current landscape of hacking exercises, their efficacy, and challenges through a set of pedagogical principles.} 
{Moreover, in an effort to promote awareness and bolster the security of AI-Robotics in academic environments, researchers \cite{mittal2023ai} are advocating for the inclusion of robotics curricula that equip students with the knowledge and abilities required for developing resilient and secure robotic applications.}


{Similarly, researchers also recommend game-based teaching approaches to make learning enjoyable, and such studies indicate that it has positive cognitive and affective effects in training people} \cite{yasin2018design, 8194898}. In their work \cite{yasin2018design}, 
{the authors designed security requirements educational games to educate stakeholders about cyber security awareness and to empirically evaluate the impact of game-based training. The game included puzzle cards with security-related terms and concepts that gave a deeper understanding of an attack in a real scenario. Along the same lines, the researchers in} \cite{8194898} 
{investigated contrasting security decisions of security experts, managers, and computer scientists from the three demographics to learn the difference in their perception of security. Their game primarily focused on player's decisions in handling physical security, firewalls, antivirus, network monitoring, and intrusion detection.}

{On the other hand, there are companies and professional associations that provide training and certification programs to educate stakeholders with the necessary skills and knowledge to secure robotic systems. For example, the Association for Advancing Automation (A3) \cite{a3} offers certification in robotics systems security to ensure that engineers and integrators have the appropriate cybersecurity knowledge and skills.}

\section{Conclusion}
\label{section_conclusion}


Today, AI-Robotics systems are deeply embedded in our everyday lives (e.g., robotic vacuum cleaners, cars with semi-autonomous capabilities, delivery robots). A common AI-Robotics system may use its sensors to collect data about itself and the environment. The physical system, algorithms, and data are all susceptible to security attacks.

In this paper, we provide an overview of attack surfaces and associated defensive strategies relevant to AI-Robotics systems.
We introduce a taxonomy of security attacks on these robotics systems and identified two major categories: physical and digital attacks. In order to highlight the vulnerabilities of AI-Robotics systems, we decompose fundamental robotic AI architectures into primitives (i.e., sense, plan, attack) and stacks (i.e., perception, navigation \& planning, and control) providing security risks at each layer. Emerging technologies, including human-robot interaction, are discussed while making note of privacy, ethical, and trustworthiness concerns. Finally, we provide our perceived future research direction to aid researchers interested in this exciting field.

\section*{Acknowledgements}
The work reported herein was supported by the National Science Foundation (NSF) (Award \#2246920). Any opinions, findings, and conclusions or recommendations expressed in this material are those of the authors and do not necessarily reflect the views of the NSF. We would like to acknowledge Charles Raines for his help in compiling literature related to ROS, OpenCV, and Tensorflow.

The authors would also like to thank the PATENT Lab (Predictive Analytics and TEchnology iNTegration Laboratory) at the Department of Computer Science and Engineering, Mississippi State University.








\ifCLASSOPTIONcaptionsoff
  \newpage
\fi

\bibliographystyle{unsrt}
\bibliography{refs}

\EOD

\end{document}